%% file: main.tex
\documentclass[a4paper,fleqn, 11pt]{cas-sc}

\usepackage{mathtools}
\usepackage{graphicx}
\usepackage[font=small,labelfont=bf]{caption}
\usepackage{subcaption}
\usepackage[sort&compress,numbers]{natbib}
\setcitestyle{numbers,square,comma,sort&compress}
\usepackage{setspace}
\usepackage{siunitx}
\usepackage{float}
\usepackage{csquotes}
\usepackage[most]{tcolorbox}
\usepackage{placeins}

\graphicspath{ {./} }

\begin{document}

\input{title}
\input{intro}
\input{review}

\input{method}
\input{results}
\input{discussion}
\input{conclusion}
\input{ack}

\newpage

\bibliographystyle{elsarticle-num-names}
\bibliography{references}

\end{document}

%% file: title.tex
\shorttitle{Generative AI-assisted Participatory Modeling in Socio-Environmental Planning under Deep Uncertainty}
\shortauthors{Zhihao Pei et~al.}

\title[mode = title]{Generative AI-assisted Participatory Modeling in Socio-Environmental Planning under Deep Uncertainty}

\tnotemark[1]

\tnotetext[1]{This research did not receive any specific grant from funding agencies in the public, commercial, or not-for-profit sectors.}

\author[1]{Zhihao Pei}[orcid=0000-0001-6609-751X]
\cormark[1] 
\ead{zppei@student.unimelb.edu.au}

\address[1]{School of Computing and Information Systems, Faculty of Engineering and Information Technology, The University of Melbourne, Parkville VIC 3010, Australia}

\address[2]{The Commonwealth Scientific and Industrial Research Organization (CSIRO), Research Way, Clayton VIC 3168, Australia}

\address[3]{Department of Psychiatry, The University of Melbourne, Parkville VIC 3010, Australia}

\author[1]{Nir Lipovetzky}
\ead{nir.lipovetzky@unimelb.edu.au}

\author[2]{Angela M. Rojas-Arevalo}
\ead{angela.rojas@csiro.au} 

\author[1,3]{Fjalar J. {de Haan}}
\ead{fjalar.dehaan@unimelb.edu.au}

\author[2]{Enayat A. Moallemi}
\ead{enayat.moallemi@csiro.au}

\cortext[cor1]{Corresponding author. Mobile: +61 421 206 369; Postal address: The University of Melbourne, Parkville VIC 3010, Australia.} 

\begin{abstract}
Socio-environmental planning under deep uncertainty requires researchers to identify and conceptualize problems before exploring policies and deploying plans. In practice and model-based planning approaches, this problem conceptualization process often relies on participatory modeling to translate stakeholders' natural-language descriptions into a quantitative model, making this process complex and time-consuming. To facilitate this process, we propose a templated workflow that uses large language models for an initial conceptualization process. During the workflow, researchers can use large language models to identify the essential model components from stakeholders' intuitive problem descriptions, explore their diverse perspectives approaching the problem, assemble these components into a unified model, and eventually implement the model in Python through iterative communication. These results will facilitate the subsequent socio-environmental planning under deep uncertainty steps. Using ChatGPT 5.2 Instant, we demonstrated this workflow on the lake problem and an electricity market problem, both of which demonstrate socio-environmental planning problems. In both cases, acceptable outputs were obtained after a few iterations with human validation and refinement. These experiments indicated that large language models can serve as an effective tool for facilitating participatory modeling in the problem conceptualization process in socio-environmental planning.
\end{abstract}
 
 
\maketitle

%% file: intro.tex
\section{Introduction} \label{intro}
Socio-environmental planning often involves complex uncertainties, ranging from well-characterized ones to deep ones. While the former are relatively tractable, the latter cannot be fully eliminated until future conditions unfold \citep{walker2013adapt}. This paper concentrates on deep uncertainty, as socio-environmental planning problems characterized by deep uncertainty are common in the real world \citep{hamarat2013adaptive, kasprzyk2013many, auping2015societal, groves2015developing, Marchau2019, workman2020decision}. According to the definition, deep uncertainty arises when people cannot determine `(1) the appropriate models to describe the interactions among a system’s variables, (2) the probability distributions to represent uncertainty about key variables and parameters in the models, and/or (3) how to value the desirability of alternative outcomes' \citep{lempert2003shaping}. These dimensions are hereafter referred to as \emph{model uncertainty}, \emph{parameter uncertainty}, and \emph{objective uncertainty}, respectively \citep{geffner2013concise}.

\begin{figure}
    \centering
    \begin{subfigure}{.45\textwidth}
        \centering
        \captionsetup{width=.9\linewidth}
        \includegraphics[width=\linewidth]{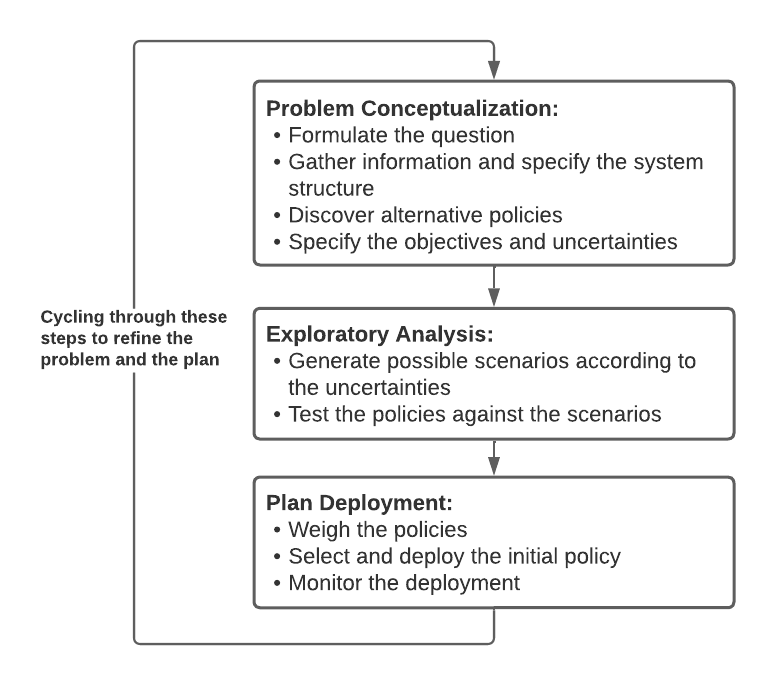}
        \caption{The general process of DMDU approaches, adapted from \citep{Marchau2019}.}
        \label{fig:DMDU}
    \end{subfigure}
    \begin{subfigure}{.45\textwidth}
        \centering
        \captionsetup{width=.9\linewidth}
        \includegraphics[width=\linewidth]{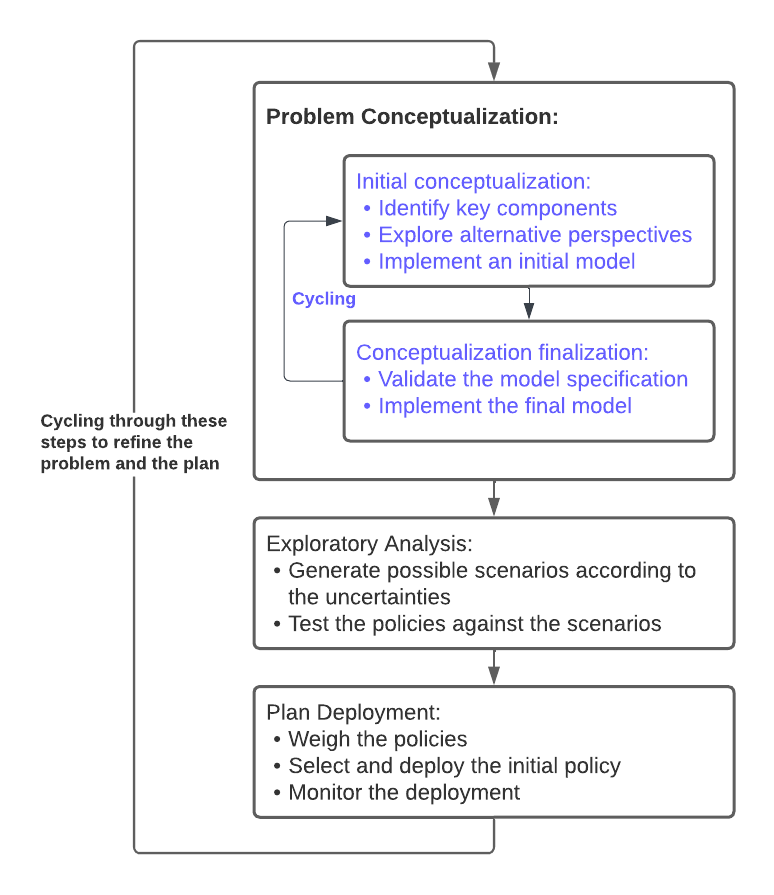}
        \caption{The updated process of DMDU approaches, supported by LLMs for initial problem conceptualization. The blue text highlights its differences from the original process.}
        \label{fig:DMDU_after}
    \end{subfigure}
    \caption{The two DMDU processes covered in this paper.}
\end{figure}

Walker et al. \citep{walker2013adapt} argued that a robust plan for Decision-Making under Deep Uncertainty (DMDU) should not only (1) achieve its objectives; but also (2) perform satisfactorily across diverse plausible futures (robust); and (3) remain adaptable as conditions evolve (adaptive). To support such planning in socio-environmental problems under deep uncertainty, several DMDU approaches have been developed \citep{lempert2007managing, groves2007new, haasnoot2013dynamic, ben2006info, korteling2013using}. These approaches generally follow a process consisting of three iterative steps (Figure \ref{fig:DMDU}): (1) problem conceptualization, (2) exploratory analysis, and (3) plan deployment \citep{haasnoot2013dynamic, Marchau2019}. First, researchers collaborate with decision-makers and other stakeholders to formalize the problem. Stakeholders here refer to those who make decisions or are affected by the decisions \citep{reed2009s}. This engagement of stakeholders in the modeling process is commonly referred to as participatory modeling \citep{halbe2020participatory, smetschka2020co}. Through this process, researchers gather and organize the essential information for decision-making into a framework or model to support the subsequent exploration. This information includes the problem context, system structure, alternative policies, objectives, and associated uncertainties. Second, candidate policies are explored across scenarios emerging from the uncertainties, in order to inform the development of the final plan. Finally, an initial plan is implemented accordingly and monitored, allowing for adaptation as the future unfolds. In practice, these three steps are typically executed iteratively, enabling researchers and decision-makers to refine the process in response to evolving conditions and thereby enhance the adaptability of the final plan.

However, it is often complex and time-consuming to translate stakeholders' understandings of the problem into a quantitative model \citep{campuzano2025natural} in participatory modeling during problem conceptualization, especially when non-expert stakeholders, such as citizens and community members, are involved \citep{voinov2010modelling}. Generally, these stakeholders describe their understandings in natural language, which can be ambiguous and may not guarantee the completeness and consistency required for formal model specification \citep{huang2025planning}. Therefore, traditional methods are laborious and require substantial manual effort for interpretation and analysis \citep{campuzano2025natural, necula2024systematic}. Moreover, different stakeholders may have different perceptions, backgrounds, and interests, thus approaching the same problem from different perspectives \citep{cvitanovic2015improving}. Building a model that can embed all these perspectives is a challenging task. This challenge has also been emphasized in government-facing contexts where planners must integrate diverse sectoral perspectives to inform planning \citep{bandari2022prioritising}.

The emergence of generative AI approaches such as Large Language Models (LLMs) marks a recent milestone in machine learning \citep{gao2024large}. These LLMs are trained on vast corpora of data, and demonstrate exceptional capability in understanding and generating natural language \citep{brown2020language, ali2024llms, chang2024survey}. Nowadays, LLMs have proven effective across a wide range of applications \citep{zhao2023survey}, such as text generation and summarization \citep{, waller2024questionable, nagano2025llm, van2023clinical, zhang2024benchmarking}, translation \citep{kocmi2023large, zhang2023prompting}, sentiment analysis \citep{alex2021raft, chen2023robust, qin2023chatgpt}, and code generation \citep{du2024evaluating, lei2025planning}. More importantly, the application of LLMs for conceptual and computational modeling is gaining increasing attention \citep{ali2024llms, gao2024large, storey2025large, aghzal2025survey, huang2025chasing}. LLMs have also been used to enhance ideation by disrupting habitual thought patterns and enabling exploration of expansive problem spaces \citep{li2025review}. Therefore, we posit that LLMs are well suited to support participatory modeling in socio-environmental planning under deep uncertainty, particularly during the early stage of problem conceptualization, when stakeholders’ understandings must be translated into structured model components.

We also recognize that a limitation of LLMs is that their outputs cannot guarantee the reliability in long-horizon planning or multi-step reasoning \citep{momennejad2023evaluating, pallagani2023understanding, valmeekam2024llms}, so LLMs are more suitable as decision aids rather than stand-alone planners \citep{aghzal2025survey}. A limitation of LLM-assisted modeling studies is that they mainly focus on the performance of produced models, with little attention to guiding users in how to interact with LLMs \citep{ali2024llms}. Accordingly, this paper proposes a \emph{templated, LLM-assisted workflow} that guides researchers in performing initial problem conceptualization prior to the conventional conceptualization. By following this workflow, they would be able to identify necessary model components, such as state variables, action options, system structures, objectives and uncertainties, based on stakeholders' intuitive ideas of the problem. They would also be able to explore diverse perspectives on the problem and identify corresponding factors influencing the system, which might otherwise be overlooked due to individual perceptions. As the output of the workflow, LLMs would implement all identified components into an initial Python program. In this way, the workflow functions as a bootstrapping and rapid-prototyping tool for researchers to facilitate the problem conceptualization process, while reducing barriers to participation for non-expert stakeholders. Contributions of this paper are summarized as follows:
\begin{itemize}
    \item This paper proposes an LLM-assisted workflow that guides researchers through the initial problem conceptualization process. This workflow improves the efficiency of problem conceptualization and reduces barriers to participation for non-expert stakeholders in socio-environmental planning under deep uncertainty, as framed by the DMDU approaches. The process of DMDU approaches with the LLM-assisted initial conceptualization is illustrated in Figure \ref{fig:DMDU_after}.
    \item This paper demonstrates the workflow on two hypothetical socio-environmental planning under deep uncertainty problems, the lake problem and an electricity market problem. The former concerns the control of annual pollutant emissions into a nearby lake, in order to balance economic benefits with pollutant concentrations in the lake. This problem is a canonical benchmark with a standard implementation publicly available. The latter concerns the bidding strategy of a wind-power producer facing production uncertainty due to socio-environmental factors in a competitive spot market, with the aim of evaluating its expected revenue. This problem has been implemented differently in previous studies, and we used it to test conditions that go against the training data used to create the LLM. In both cases, the LLM successfully identified model components and alternative perspectives consistent with the baseline problem descriptions, and implemented them in Python in a syntactically and logically correct manner after only a few iterations of human validation and refinement. 
    \item At the end, this paper discusses the strengths and weaknesses for applying the workflow to facilitate participatory modeling during problem conceptualization in socio-environmental planning under deep uncertainty, based on the literature review and experiments conducted.    
    \item This research also opens up new frontiers for socio-environmental planning under deep uncertainty and for DMDU more broadly by introducing LLMs as effective tools to facilitate the decision-making process. It also reviews related applications of LLMs for modeling and planning across multiple fields, highlighting an emerging trend in decision-making.
\end{itemize}

The remainder of this paper is organized as follows. Section 2 reviews background and related work on DMDU and LLMs. Section 3 introduces the proposed LLM-assisted workflow. Section 4 details the experimental design, including the case studies and evaluation methods. Section 5 presents the experimental results. Section 6 provides a balanced assessment of the strengths and limitations of applying the workflow for initial problem conceptualization in socio-environmental planning, and Section 7 concludes this paper. 

%% file: review.tex
\section{Literature Review}
\subsection{Problem conceptualization in DMDU}
Common DMDU approaches include \emph{Robust Decision-Making (RDM)} \citep{lempert2007managing, groves2007new}, \emph{Dynamic Adaptive Policy Pathways (DAPP)} \citep{haasnoot2013dynamic}, and \emph{Info-Gap Decision Theory (IGDT)} \citep{ben2006info, korteling2013using}. All these approaches begin with the problem conceptualization step that translates stakeholders' knowledge of the problem into model specification before subsequent steps can proceed. Across these approaches, high-quality conceptualization is a prerequisite that enables efficient and effective exploratory analysis and robust plan development \citep{Marchau2019}. In the problem conceptualization process of \emph{RDM}, stakeholders are involved in identifying the key factors of the problem, including the objectives, action options, uncertainties, and the relationships between them \citep{lempert2019robust}. This information is then organized into an XLRM framework \citep{lempert2003shaping}. The problem conceptualization process in \emph{DAPP} comprises three steps: (1) describe the study area; (2) analyze the problem, vulnerabilities and opportunities; and (3) identify possible actions \citep{haasnoot2013dynamic}. \emph{IGDT} identifies three components through problem conceptualization: the model, performance requirements, and uncertainty model \citep{ben2019info}. Review of these frameworks showed that these studies specify which components are required but offer little guidance on how to perform the problem conceptualization process more efficiently. 

\subsection{Participatory modeling}
Collaboration with stakeholders in modeling and model-based decision-making has a long history in DMDU and socio-environmental planning research \citep{moallemi2023knowledge}. It is sometimes framed as, or associated with participatory modeling. Participatory modeling is a powerful approach that can not only deepen stakeholders' understanding of the system, but also clarify the impacts of solutions to the decision-making problem \citep{voinov2010modelling}. The literature has documented a sustained increase in attention to participatory modeling in recent years \citep{voinov2016modelling, moallemi2021evaluating}. Generally, the participatory modeling process consists of seven components: (1) \emph{scoping \& abstraction}, (2) \emph{envisioning \& goal-setting}, (3) \emph{model formulation}, (4) \emph{data, facts, logic, cross-checking}, (5) \emph{model application to decision-making}, (6) \emph{evaluation of outputs \& outcomes}, and (7) \emph{facilitation of transparency}. It is worth noting that the level of stakeholder involvement in different components in a project depends on the particulars of the participatory modeling process. Not all modeling processes require stakeholder involvement in every component, and any particular stakeholder is unlikely to participate in all components. Additionally, Voinov et al. \citep{voinov2016modelling} reviewed recent tools and studies associated with each component. The LLM-assisted initial problem conceptualization introduced in this paper is associated with the first three components.

\subsection{LLMs for conceptual modeling}
As we mentioned, research on LLM for conceptual modeling is gaining increasing attention recently \citep{arulmohan2023extracting, camara2023assessment, chaaben2023towards, chen2023use, chen2023automated, fill2023conceptual}. These studies mainly examined the quality of models produced by LLMs. In contrast, Ali et al. \citep{ali2024llms} investigated modelers’ interaction patterns with LLMs, as well as their perceived usefulness and ease of use in this process. In their study, an empirical experiment was conducted, in which 76 students were asked to create conceptual models with one of two LLMs in one of three application domains. The results were summarized in an interaction process model. This model included the recurrent intentions employed by the students and the probability distributions of transitions between the intentions. This model captured the typical interaction patterns of how students interacted with LLMs in conceptual modeling. The results also revealed several differences in terms of interaction patterns across different LLMs and different application domains. As for perceived usefulness and ease of use of LLMs for conceptual modeling, the students reported a moderate user experience. On the one hand, LLMs were responsive, efficient, and creative, and provided guidance on students' tasks. On the other hand, LLMs exhibited weaknesses in continuity, accuracy, reliability, visualization, and were sensitive to user inputs. These weaknesses have also been documented in related studies, and further improvements to LLMs were encouraged to enhance performance in conceptual modeling \citep{camara2023assessment, storey2025large}. For example, one possible direction is to develop prompt templates that guide user–LLM interactions \citep{ali2024llms}, which is one of the contributions of this paper.

\subsection{LLMs for automated planning} \label{llm_for_planning}
LLMs have also been widely used in automated planning, a branch of AI, with a long tradition \citep{newell1959report}, concerned with the study of computational models and planners to provide agents with reasoning and planning capabilities \citep{ghallab2016automated, geffner2013concise}. Here, a planner refers to the algorithm or software system that automatically constructs a plan to achieve the objective. Based on the role of LLMs for automated planning, their application can be classified into two groups: \emph{LLMs as stand-alone planners} and \emph{LLMs integrated with traditional planners} \citep{aghzal2025survey, tantakoun2025llms, pallagani2024prospects}.

\emph{LLMs as stand-alone planners} means that LLMs are used to propose plans directly \citep{huang2025chasing}. These approaches can be further classified into four subgroups \citep{aghzal2025survey}. \emph{Hierarchical task breakdown} decomposes a complex planning problem into simpler, manageable subproblems, and uses LLMs to solve them incrementally. Nevertheless, these approaches often rely on problem-specific prompts, limiting their generality to novel problems, and fail to scale up to large problems. They are also limited to applications where the system states are fully observable to enable the decomposition. \emph{Plan refinement} uses LLMs to iteratively improve plans based on external or internal feedback. These approaches offer flexibility in stochastic and partially observable applications, but suffer from high computational cost and unstable feedback gains. \emph{Search-based} approaches search over candidate paths of reasoning, where LLMs are used for exploration and evaluation in the search space. These approaches also suffer from poor reliability of LLMs and high cost in complex applications. \emph{Fine-tuning} exposes LLMs to planning instances and their solutions to improve plan generation over similar problems, but these plans generalize poorly. 

The limitations of previous approaches motivated \emph{LLMs integrated with traditional planners}, where LLMs are used as auxiliary components in planning. These approaches can also be classified into three subgroups \citep{aghzal2025survey}. \emph{Text-to-formal language translation} uses LLMs to convert natural-language task descriptions into formal languages \citep{huang2025planning}, such as Planning Domain Definition Language (PDDL) \citep{mcdermott1998pddl, haslum2019introduction}, which traditional planners, such as Best-First Width Search (BFWS) \citep{lipovetzky2017best}, can process to propose plans. By combining the natural-language processing capabilities of LLMs with the efficiency and rigor of traditional planners, these methods make planning more accessible to non-expert users. \emph{Knowledge augmentation} integrates the common-sense knowledge of LLMs acquired from training on diverse and extensive corpora into planning. This provides semantic cues to guide planning and reduces the need for manual input of domain knowledge. \emph{Plan evaluation} positions LLMs as critics to evaluate plans produced by traditional planners. These approaches can make planning more versatile, since the common-sense knowledge of LLMs enables them to capture some important subtleties in natural-language descriptions that may be overlooked in reward function designs. The ideas of these approaches motivated our use of LLMs for initial problem conceptualization, including translating intuitive problem descriptions into model specifications and exploring diverse perspectives and factors of the problem. 

\subsection{Prompt engineering} \label{techniques}
In the context of LLM, a prompt is a task-specific instruction given to an LLM to generate a desired response without modifying the LLM \citep{sahoo2024systematic}. Prompt engineering is an emerging discipline that designs and optimizes such prompts to improve the quality of LLM responses, in order to enhance the capabilities of LLMs across various applications \citep{chen2025unleashing}. In addition, prompt engineering also helps researchers better understand LLMs in structured reasoning and generative tasks. To improve the performance of LLMs in problem conceptualization, we used the following two prompt engineering techniques in our workflow design:
\begin{enumerate}
    \item \emph{Chain-of-thought prompting} \citep{wei2022chain} is a technique that helps LLMs solve complex, multi-step reasoning tasks by explicitly providing intermediate reasoning steps, instead of jumping directly to the final answer.
    \item \emph{Prompt chaining} \citep{sahoo2024systematic} is a technique that uses multiple prompts sequentially to accomplish a complex task that cannot be effectively handled in a single prompt. Each prompt builds on the output of the previous one, forming a chain of interactions that progressively refines reasoning, generates structured outputs, or performs multi‑stage workflows.
\end{enumerate}
For a broader overview of prompt techniques, we refer readers to \citep{sahoo2024systematic}.

%% file: method.tex
\section{LLM-assisted Workflow}
\begin{figure}
    \centering
    \captionsetup{width=.9\linewidth}
    \includegraphics[width=\linewidth]{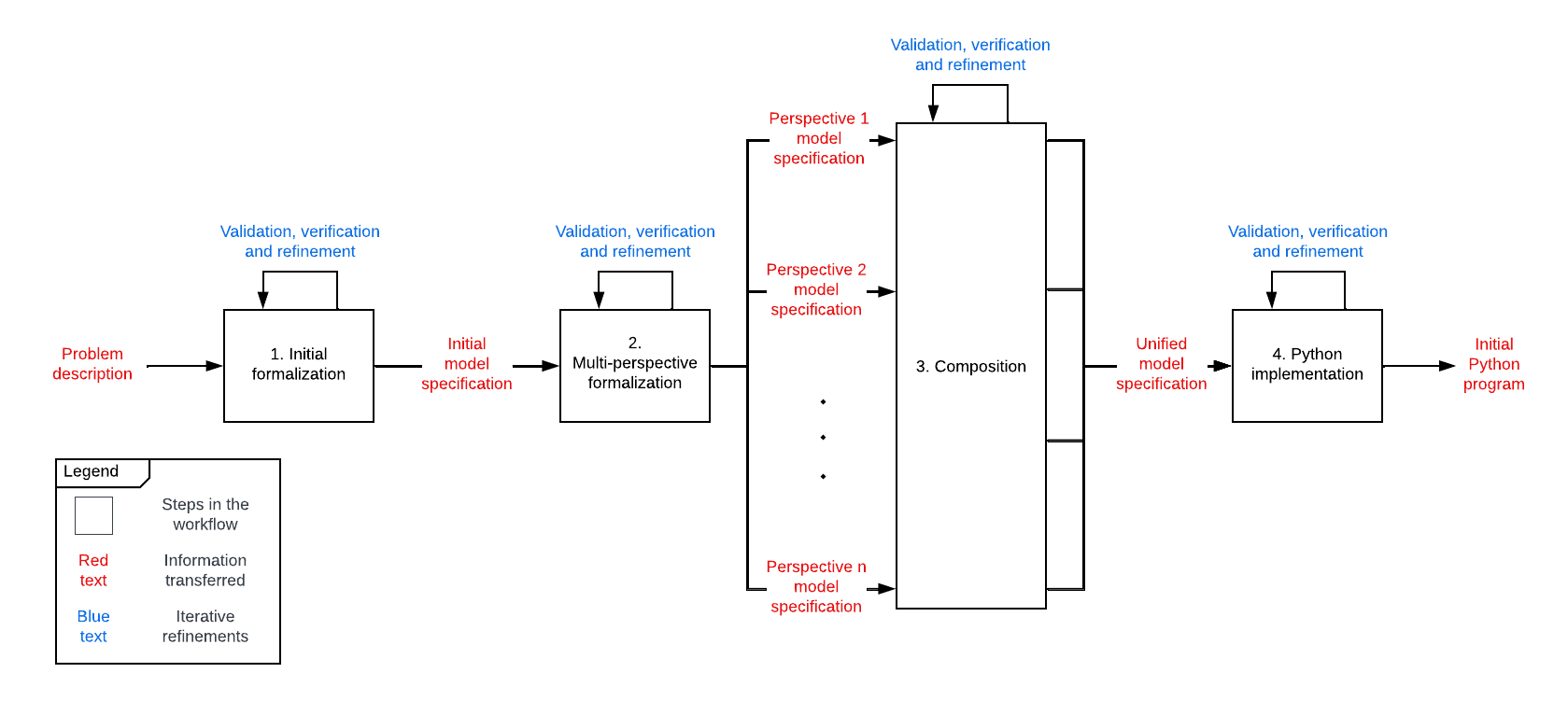}
    \caption{The LLM-assisted workflow for communication with LLMs for initial problem conceptualization. In this diagram, nodes represent discrete steps in the workflow. Red text denotes the information transferred between steps, whereas blue text denotes iterative human validation, verification and refinement of the responses.}
    \label{fig:template}
\end{figure}

This section presents the proposed LLM-assisted workflow that guides researchers, decision-makers and other stakeholders in communicating with LLMs to perform initial problem conceptualization (Figure \ref{fig:template}). This workflow starts with a prompt containing stakeholders' intuitive problem description. The information is then processed through four steps. At each step, researchers collaborate with stakeholders to engage in interactive dialogue with LLMs to validate and verify outputs and iteratively refine prompts until satisfactory results are obtained. The final outputs consist of a multi-perspective model specification of the problem, along with its Python implementation that functions as a computational simulation model. This workflow is suitable for model-based implementation of various DMDU approaches. Specifically, the multi-perspective model specification supports conceptual formalization of the decision-making problem. The Python implementation is compatible with Rhodium \citep{hadjimichael2020rhodium} and the EMA Workbench \citep{kwakkel2017exploratory}, which are two popular implementation packages for model-based DMDU approaches, thereby enabling computational exploration and analysis in subsequent steps. Note that the prompts were crafted in a specific way in our experiments so that the final Python implementation could be directly used in EMA Workbench. With appropriate modifications to these prompts, the Python implementation can also be adapted for other packages.

\subsection{Design logic}
We adopted \emph{chain-of-thought prompting} and \emph{prompt chaining} (Section \ref{techniques}) as proven prompt engineering techniques to structure this workflow. First, we decompose the complex task of initial problem conceptualization into four intermediate reasoning steps. This decomposition allows the LLM to address simpler sub-tasks incrementally, which collectively compose the final outputs. This decomposition mitigates the limitation of the LLM in handling long-term reasoning. Second, we explicitly specify the outputs of each step, which serve as the inputs for subsequent steps. This technique establishes a chain of interactions that enhances the LLM’s effective memory and improves the consistency of its outputs. Note that our experiments only evaluated the overall workflow. The assessment of the causal contributions of individual techniques is left for future work.

\subsection{Initial formalization}
Step 1 is initial formalization. At this stage, researchers' prompt to the LLM should contain two parts: (1) stakeholders' intuitive problem description; and (2) instructions that prompt the LLM to formalize the problem and identify key components for modeling. For (1), a \emph{comprehensive} problem description should contain sufficient information to allow identification of all components necessary for model specification, such as our problem descriptions in Section \ref{problems}. This description can only be given in a well-defined and known problem, or after sufficient communication among researchers, decision-makers and other stakeholders, which is rarely the case in real applications. In fact, it is acceptable for stakeholders to provide a \emph{brief}, narrative-based, and intuitive problem description, where necessary components may be omitted. This gap can be addressed by (a) having the LLM augment the description using its common-sense knowledge \citep{aghzal2025survey}; and (b) iteratively refining the responses through human validation and verification. This process not only improves the quality of model specification, but also helps researchers gain deeper and more structured understandings of their problem. We demonstrated both \emph{comprehensive} and \emph{brief} descriptions in our experiments.

In the experiments, we used the following instruction for (2):
\begin{displayquote}
\emph{Please formalize this problem description as a simulation model and specify the key components, including state variables, decision variables, transition functions, objective functions, stochastic variables, constant parameters and deep uncertainties.}
\end{displayquote}
We prompted the LLM to formalize the problem as a simulation model without specifying a specific model type. This is because the exploration function in EMA Workbench is likewise not restricted to any specific model type. In fact, this function only requires a simulation model that takes a scenario and decisions as inputs and returns objective values as outputs. We also prompted the LLM to explicitly identify the key components covered in the problem description. Here, state variables are variables that characterize the system’s states. Decision variables are controllable choices adjusted by decision-makers to achieve the objectives. Transition functions describe how the system transitions from one state to the next. Objective functions reflect the objectives of the problem. In addition, stochastic variables, constant parameters and deep uncertainties are components that make up the transition functions and objective functions. In real applications, researchers can also specify the target model type as needed, such as Markov decision process \citep{puterman1990markov, rl2018mdp} and agent-based model \citep{gilbert2019agent}, and specify the corresponding model components in the prompt. The expected output of Step 1 is a model specification with the specified components. This specification is formed based on the input problem description and the common-sense knowledge of the LLM.

Explicitly specifying the required components offers several advantages. First, it encourages the LLM to explore these components based on its common-sense knowledge, thereby complementing researchers' and stakeholders' understandings of the problem. Second, it supports the adoption of \emph{prompt chaining} in the workflow. 

\subsection{Multi-perspective formalization}
In real-world socio-environmental planning, the system can often be complex and involve multiple perspectives representing diverse stakeholder groups with different perceptions, backgrounds, and interests. For example, the lake problem \citep{carpenter1999management, singh2015many} can be approached from the perspectives of government, individuals, and industrial owners. By accounting for all these perspectives in the problem conceptualization process, researchers can develop a comprehensive model specification that captures relevant factors arising from stakeholder actions and the impacts on the stakeholders. Such a specification will support the development of appropriate final plans. Therefore, the LLM is prompted to infer multiple perspectives approaching the original problem, and to formalize them into different model specifications in Step 2. Each perspective might be associated with different decision variables, transition functions and objectives. The instruction we used is:
\begin{displayquote}
\emph{Various stakeholders approach this problem from different perspectives. Please identify different perspectives related to this problem. These perspectives should share a common environment, with each stakeholder independently controlling its own decision variables. In the specification, specify the common environment, including the global parameters and variables. For each perspective, also specify the corresponding model and its key components, including the decision variables, transition functions, and objective functions. If there are multiple objectives, provide multiple objective functions instead of a weighted objective function.}
\end{displayquote}
The expected output of Step 2 consists of two parts: (a) the environment shared across all perspectives, including the global parameters and variables; and (b) multiple model specifications specific to each perspective, including the decision variables, transition functions, and objective functions. Note that we specified several requirements in this prompt, including (1) identifying multiple perspectives without explicitly naming them; (2) ensuring that the perspectives independently influence the common environment; and (3) specifying multiple objective functions instead of aggregating them into a weighted-sum. These requirements, as well as those specified in subsequent prompts, are optional. They were used solely to structure the output model specifications and Python implementations, thereby ensuring their comparability. Researchers may specify their own requirements as needed. Eventually, researchers should select the perspectives they find relevant, and iteratively interact with the LLM to validate and refine the corresponding specifications.

As discussed before, researchers may overlook important subtleties in problem conceptualization. Therefore, Step 2 of the workflow employs the LLM to support researchers in brainstorming different perspectives, and identifying possible factors relevant to the problem. This process helps researchers form a more comprehensive understanding of the problem and improve the quality of model specification. It also helps to explore possible model and objective uncertainties involved in the problem. 

\subsection{Composition}
In Step 3, the LLM is prompted to compose the perspectives selected in Step 2 into a single unified model. The instruction is:
\begin{displayquote}
\emph{To ensure internal consistency across these models, please compose them into a single, unified model that shares a common environment and states, while embedding perspective-specific decisions and objectives. The composed model should: 1. embed each perspective’s decision variables and objectives; 2. enable modular evaluation of each perspective; 3. for each perspective, treat other perspectives’ decisions as exogenous inputs; and 4. incorporate all other components from every perspective.}
\end{displayquote}
The unified model inherits the common environment from Step 2, while embedding the perspective-specific components. To facilitate this composition and prepare the model for modular implementation in Python, we include four guiding principles in the prompt to structure the response. The expected output is a unified model specification consisting of two parts: (a) the common environment carried over from Step 2; and (b) modular specifications of perspective-specific components for each perspective. In each modular specification, decision variables from other perspectives are treated as exogenous inputs. For example, for the government perspective on the lake problem, the decision variables for individuals and industrial owners are treated as exogenous inputs. 

This step serves two purposes: (1) this step enables the unified model to simulate the system simultaneously influenced by the decisions of multiple stakeholders, which is common in real applications; and (2) composing multiple perspectives into a unified model requires the LLM to explicitly describe the interactions among these perspectives. This explicit representation ensures consistency across the perspectives, thereby enabling their comparability within a single model.

\subsection{Python implementation}
Eventually, the LLM is prompted to implement the unified model in Python in Step 4. The instruction is:
\begin{displayquote}
\emph{Provide a modular Python implementation of the unified model, using classes to represent different perspectives. At the end of the program, define a function as the interface. This function should take as inputs a dictionary of decision variable values, a dictionary of uncertain parameter values, and a dictionary of constant parameter values; simulate the model dynamics; and return the value of each objective function. This function does not need to be executed.}
\end{displayquote}
In the output Python program, the common environment and each perspective are translated into a distinct Python class, allowing researchers to simulate every component independently by invoking the corresponding class. This structure supports modularity and flexibility in exploration. The implementation is required to accept the values of decision variables and uncertain parameters as inputs, simulate the dynamics of the model, and return the value of each objective function. In this way, the final program can be directly used for exploratory analysis in EMA Workbench.

\subsection{Validation, verification and refinement}
At each step of the workflow, researchers need to iteratively interact with the LLM to validate, verify and refine its responses until satisfactory results are obtained. Here, validation refers to checking whether the output model specification is consistent with the input problem description, while verification refers to demonstrating that the output Python implementation works as intended through testing \cite{roache1998verification}. This process can be implemented in various ways \citep{sargent2010verification}. Owing to the advanced natural language processing capabilities of the LLM, researchers should be able to conduct this process through conversational interactions analogous to engaging with a human interlocutor. As one concrete instantiation, our workflow employs a \emph{checklist} to validate the outputs of Step 1–3, and \emph{testing} to verify the outputs of Step 4.

\subsubsection{Checklist} \label{checklist}
Our checklist is shown below, which is inspired by the \emph{Overview, Design concepts and Details (ODD)} protocol \citep{grimm2010odd, grimm2020odd}. This protocol has been widely accepted for describing various types of simulation models. For Step 1–3, researchers should sequentially check whether each item is consistent with their understandings of the problem. If not consistent, they should issue a refinement prompt to the LLM specifying the discrepancy.
\begin{enumerate}
    \item \emph{Purpose, scope and patterns:} Validate whether the model specification produced by the LLM accurately captures the purpose and scope of the problem, as well as the patterns in the system. 
    \item \emph{Entities and state variables:} Validate whether the specification involves correct entities and characterizes system states with appropriate variables. 
    \item \emph{Design concepts:} Validate model design concepts specific to the model type used. For our simulation models, these concepts include the objective functions, stochastic variables, constant parameters, and uncertainties.
    \item \emph{Processes and scheduling:} Validate whether the processes that describe system dynamics, and their scheduling, are consistent with researchers’ understandings of the problem.
    \item \emph{Perspectives:} Validate whether the multi-perspective specification proposed by the LLM captures stakeholders’ interests and remains mutually consistent within the common environment. 
\end{enumerate}
Note that our checklist does not include the \emph{Details} components from the original ODD protocol, which include
\begin{enumerate}
    \item \emph{Initialization:} Describes the initial state of the model.
    \item \emph{Input data:} Describes the external data inputs to the model.
    \item \emph{Submodels:} Describe the full details of the processes in \emph{processes and scheduling}.
\end{enumerate}
This is because the first two, \emph{initialization} and \emph{input data}, will be validated through testing in Step 4 of the workflow, while \emph{submodels} can also be validated in the \emph{processes and scheduling} process for simplicity. For this reason, we also place \emph{processes and scheduling} after \emph{design concepts} in our checklist.

\subsubsection{Testing} \label{testing}
To guarantee the syntactic and logical correctness of the final Python implementation, our workflow employs the following three tests for verification in Step 4 of the workflow. 
\begin{enumerate}
    \item \emph{Unit tests} test specific functionalities in the implementation.
    \item \emph{Property-based tests} test whether the implementation conforms to the properties specified in the problem description.
    \item \emph{Scenario tests} test whether the implementation yields the expected performance in stakeholder-authorized scenarios.
\end{enumerate}
We did not include tests specifically for syntactic correctness, as any syntax errors would be exposed by the following tests through execution failure.

\section{Methodology}
Our experiments aim to demonstrate the proposed LLM-assisted workflow for guiding researchers in using LLMs to perform initial problem conceptualization in socio-environmental planning problems under deep uncertainty. To achieve this, we used ChatGPT 5.2 Instant to formalize two case studies by following the workflow and evaluated the performance. To mitigate the influence of output variability inherent to the LLM, each case study was conducted several times. This section describes the corresponding experimental design.

\subsection{ChatGPT 5.2 Instant}
In this paper, we used ChatGPT 5.2 Instant for demonstration in our experiments. ChatGPT is one of the most widely-used LLMs developed by OpenAI for general interactions \citep{gpt}. ChatGPT 5.2 has demonstrated better performance across multiple domains, such as coding, writing and math, compared to OpenAI's previous models \citep{gpt52}. Nowadays, ChatGPT 5.2 has two versions available online: ChatGPT 5.2 Instant and ChatGPT 5.2 Thinking. We chose ChatGPT 5.2 Instant over Thinking for two reasons: (1) ChatGPT 5.2 Instant was more time-efficient for our case studies; and (2) ChatGPT 5.2 Thinking consistently introduced additional components that were not specified in the input problem description into the model specification, which added unnecessary complexity.

\subsection{Case Studies} \label{problems}
\begin{figure}
    \centering
    \captionsetup{width=.9\linewidth}
    \includegraphics[width=\linewidth]{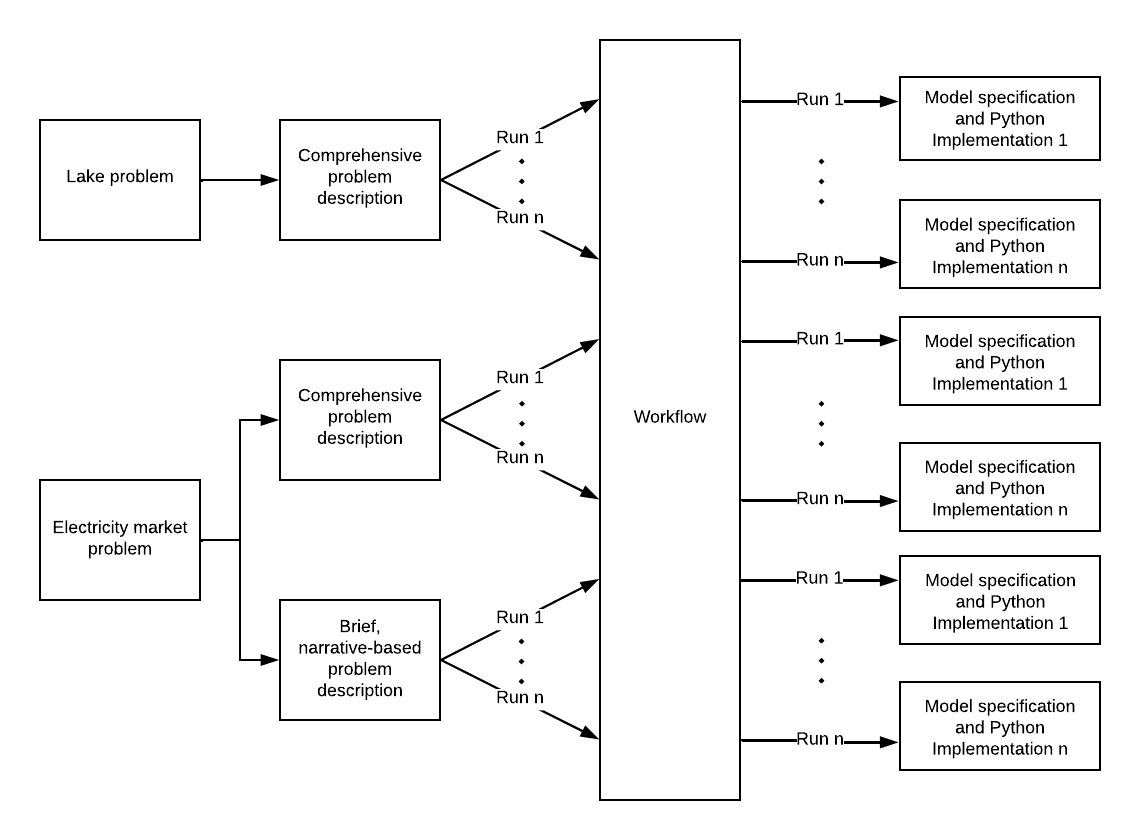}
    \caption{The overall experimental design. In our experiments, we considered two case studies, the lake problem and the electricity market problem. For the former, we only considered the sub-case with a \emph{comprehensive} problem description. For the latter, we considered both the sub-case with a \emph{comprehensive} problem description and the sub-case with a \emph{brief} problem description. For each sub-case, we ran the initial problem conceptualization process several times following the proposed workflow.}
    \label{fig:experiments}
\end{figure}

Our experiments demonstrated the workflow on two hypothetical socio-environmental planning problems under deep uncertainty framed by DMDU: the lake problem and an electricity market problem. Both of them describe situations where people make decisions that affect a societal or environmental system. The former is a canonical benchmark with a publicly available standard implementation, while the latter has been implemented differently in previous studies. The following two sections provide their \emph{comprehensive} baseline descriptions. For this reason, we always provided the \emph{comprehensive} problem description specifying all necessary components in the lake problem. In contrast, two sub-cases were considered for the electricity-market problem. In the first sub-case, the \emph{comprehensive} problem description was provided. In the second, only a \emph{brief} description that omitted many components was provided to demonstrate a situation where the communication is limited and stakeholders are less familiar with model-based planning process. These experiments evaluate the workflow under high- and low-information settings, providing a more comprehensive assessment of the workflow. In summary, Figure \ref{fig:experiments} shows the experimental design.

\subsubsection{Lake problem} \label{lake_problem}
The lake problem describes a situation where inhabitants of a town intend to increase their economic benefits through developing industry and agriculture \citep{carpenter1999management, singh2015many}. These activities will also emit pollution into a lake nearby. Although the lake pollution decreases continuously due to natural removal, once it exceeds a threshold $X_{crit}$, irreversible lake eutrophication would occur and cause huge losses. This problem requires decision-makers to decide the annual pollution emissions, with the aim of maximizing the economic benefit while avoiding eutrophication. The lake pollution transition function is given by
\begin{equation}
    X_{(t+1)}=X_t+a_t+\frac{(X_t^q)}{(1+X_t^q )}- bX_t+\epsilon_t
\end{equation}
where $X_t$ is the pollution at time $t$, and $X_0 = 0$; $a_t$ is the rate of anthropogenic pollution at time $t$; $b$ is the natural removal rate of the pollution; $q$ is the recycling exponent; $\epsilon_t$ is the rate of natural pollution at time $t$. The threshold $X_{\text{crit}}$ is the pollution level at which the lake’s natural recycling equals its natural removal rate. The total benefit from the pollution is given by
\begin{equation}
    f_{economic} = \sum\limits_{t \in {T}}\alpha a_t \delta^t
\end{equation}
where $\alpha$ is the benefit-to-pollution ratio, and $\delta$ is the discount factor.

The lake problem is characterized by parameter uncertainty. The natural pollution $\epsilon_t$ follows a log-normal distribution with mean $\mu$ and variance $\sigma^2$. In contrast, the probability distributions of the parameters $\mu$, $\sigma$, $b$, $q$, $\delta$ cannot be determined.

\subsubsection{Electricity market problem} \label{market_problem}
An energy company is interested in investing in wind-power production. To assess the viability of this investment, the company wants to evaluate the expected revenue of a wind-power producer in a competitive spot market.

Suppose there are five energy producers participating in the day-ahead spot market. Three of them use conventional sources for energy production, such as coal, while the other two producers use solar and wind power, respectively. The market is cleared once per day for 24 hourly intervals of the following day. For each dispatch interval $t=1, \dots, 24$, every producer submits a bid $(b_t, p_t)$ specifying the energy quantity $b_t$ (MWh) it is willing to supply and the corresponding minimum acceptable price $p_t$ (per MWh). Then, the market operator follows a merit-order process to determine the market-clearing price $c_t$ and dispatch schedule, ensuring the total market demand $D_t$ is met, where $D_t \sim \mathcal{N}(\mu_D, {\sigma_D}^2)$ with constant mean and variance. Merit-order ensures that any bid $(b_t, p_t)$ at interval $t$, if $p_t \leq c_t$, the bid is accepted, and the producer is committed to delivering $b_t$ at $c_t$. Otherwise, the producer earns zero revenue. 

From the perspective of the wind-power producer, the bid $(b_{it}, p_{it})$ from a conventional producer $i$ can be modeled as follows: $b_{it}$ can be assumed to be stable, while $p_{it} \sim \mathcal{N}(\mu_{pi}, {\sigma_{pi}}^2)$ with deeply uncertain mean and variance since its bidding strategy is unknown (inspired by \citep{rojas2022sustainability, jain2023multi}). For the solar-power producer, its $b_{st}$ is associated with the time $t$ and assumed to be $b_{st} = max(0, a+b\cos(\frac{2\pi t}{24}))$ (inspired by \citep{kaplanis2007model}), where $a$ and $b$ can be forecast based on historical data thereby assumed to be constants. Its $p_{st}$ is also uncertain, where $p_{st} \sim \mathcal{N}(\mu_{ps}, {\sigma_{ps}}^2)$ with deeply uncertain mean and variance. 

The actual energy production of the wind-power producer $G_t$ is also uncertain, as it depends on weather conditions. The production $G_t$ can be forecast from historical data and is modeled as $G_t \sim \mathcal{N}(\mu_G, {\sigma_G}^2)$ with constant mean and variance. In this case, the actual production may differ from the quantity committed to dispatch one day in advance. If the producer under-delivers relative to its dispatched quantity, the shortfall incurs a penalty of $q_u$ per MWh. This cost is imposed by the market operator to maintain the grid’s safe operational conditions and to ensure supply-demand balance in real time. 

The energy company aims to build a simulation model to evaluate the expected revenue of the wind-power producer on any future day by selecting the hourly bid quantity and price $(b_{wt}, p_{wt})$, while accounting for uncertain generation and stochastic market clearing prices. This model can then be used to explore outcomes across various short-term and long-term scenarios, thereby informing the viability of the investment. 

\subsection{Evaluation}
In our experiments, we assumed that the baseline problem description in Section \ref{problems} represents the problem the researchers sought to formalize, irrespective of whether the initial prompt provided a \emph{comprehensive} description or a \emph{brief} narrative. Therefore, at each step of the workflow, we used `whether the LLM response was consistent with the baseline description' as the criterion to guide the validation, verification and refinement process. A response was considered \emph{consistent} if (1) it contained all components covered in the baseline description; (2) the additional components introduced by the LLM did not conflict with that description; and (3) the final Python implementation contained no syntax or logic errors. If the response was consistent, we proceeded to the next step; otherwise, we iteratively issued refinement prompts until the response was satisfactory. Then, we evaluated our workflow regarding the consistency along the following three aspects.

\subsubsection{Components extraction consistency}
First, we evaluated the consistency in correctly extracting the key components in the baseline descriptions. The key components specified in these descriptions are enumerated in Table \ref{lake_components} and \ref{market_components}. In evaluation, we used the number of these components correctly identified by the LLM before and after refinements in Step 1 as the first metric. We did not report these numbers for Step 2-4 because all these components were correctly identified by the end of Step 1, and were preserved with minor errors in subsequent steps. This observation can be confirmed by the conversation logs.

\setlength{\tabcolsep}{3pt}
\begin{table}
  \begin{tabular*}{\tblwidth}{ >{\raggedright\arraybackslash}p{3.9cm}
                               >{\raggedright\arraybackslash}p{3cm}
                               >{\raggedright\arraybackslash}p{9cm} }
   \toprule
    Types & Parameters & Descriptions \\
   \midrule
    State variables & $X_t$ & Lake pollution at time $t$. \\
    Decision variables & $a_t$ & Rate of anthropogenic pollution at time $t$. \\
    Transition functions & $f(X_t)$ & System dynamics from $X_t$ to $X_{t+1}$. \\
    Stochastic variables & $\epsilon_t$ & Rate of natural pollution at time $t$. \\
    Constants & $\alpha$ & Benefit-to-pollution ratio. \\
    Deep uncertainties & $(\mu, \sigma)$; $b$; $q$; $\delta$ & Mean and standard deviation of the log-normal distribution for natural inflows; Natural removal rate; Recycling exponent; Discount factor. \\
    Objective functions & $f_{economic}$ & Total economic benefit from the pollution. \\
    Other components & $X_{crit}$ & Irreversible lake eutrophication threshold. \\
   \bottomrule
  \end{tabular*}
  \caption{The key components specified in the baseline description of the lake problem.}
  \label{lake_components}
\end{table}

\begin{table}
  \begin{tabular*}{\tblwidth}{ >{\raggedright\arraybackslash}p{3.9cm}
                               >{\raggedright\arraybackslash}p{3cm}
                               >{\raggedright\arraybackslash}p{9cm} }
   \toprule
    Types & Parameters & Descriptions \\
   \midrule
    Decision variables & $(b_{wt}, p_{wt})$ & Bid for the wind-power producer at time $t$. \\
    Transition functions & $h_t(\cdot)$ & Specification on how the market operator determines the market-clearing price and dispatches the schedule. \\
    Stochastic variables & $D_t$; $p_{it}$; $p_{st}$; $G_t$  & Market demand at time $t$; Bid price for the conventional producer $i$ at time $t$; Bid price for the solar-power producer $s$ at time $t$; Actual wind-power production at time $t$. \\
    Constants & $(\mu_D, \sigma_D)$; $b_i$; $(a,b)$; $(\mu_G, \sigma_G)$; $q_u$ & Mean and standard deviation of market demands; Bid quantity for the conventional producer $i$; Solar-power production parameters; Mean and standard deviation of wind-power production; Shortfall penalty coefficient. \\
    Deep uncertainties & $(\mu_{pi}, \sigma_{pi})$; $(\mu_{ps}, \sigma_{ps})$ & Mean and standard deviation of bid prices for the conventional producer $i$ ; Mean and standard deviation of bid prices for the solar-power producer $s$.\\
    Objective functions & $\Pi$ & Daily expected revenue. \\
    Other components & $b_{st}$ & Bid quantity for the solar-power producer $s$.\\
   \bottomrule
  \end{tabular*}
  \caption{The key components specified in the baseline description of the electricity market problem.}
  \label{market_components}
\end{table}

\subsubsection{Model formalization consistency}
Second, we evaluated the consistency in producing model specifications consistent with the baseline descriptions. Since we iterated until the LLM response was consistent with the baseline at each step, we used the number of iterations required for refinement as the second metric. By quantifying the additional effort required by the LLM to maintain consistency, this metric reflected the effort to reach consistency of model formalization when following the workflow.

\subsubsection{Python implementation consistency} \label{scenario}
Eventually, we evaluated the consistency in the performance of final Python implementations. In each sub-case of the case studies, we plotted the time series of several variables from the Python implementations across all experimental runs conducted under identical settings. These variables are shown in Table \ref{lake_series} and \ref{market_series}, while the settings are shown in Table \ref{lake_parameter} and \ref{market_parameter}. Theoretically, the overall performance and the interactions among different perspectives in these Python implementations should be consistent with the baseline problem description. Moreover, within the same sub-case and under identical settings, Python implementations generated across different experimental runs should also exhibit consistent performance, given the same input problem description. 

\begin{table}
  \begin{tabular*}{\tblwidth}{ >{\raggedright\arraybackslash}p{4cm}
                               >{\raggedright\arraybackslash}p{8.6cm}
                               >{\raggedright\arraybackslash}p{5cm} }
   \toprule
    Variables & Descriptions & Formulas \\
   \midrule
    Lake pollution & Lake pollution at time $t$. & $X_t$ \\
   \bottomrule
  \end{tabular*}
  \caption{Time series of variables used to evaluate the consistency in the performance of final Python implementations in the lake problem.}
  \label{lake_series}
\end{table}

\begin{table}
  \begin{tabular*}{\tblwidth}{ >{\raggedright\arraybackslash}p{3cm}
                               >{\raggedright\arraybackslash}p{9cm}
                               >{\raggedright\arraybackslash}p{4.7cm} }
   \toprule
    Variables & Descriptions & Formulas \\
   \midrule
    Clearing price & Clearing price at time $t$. & $c_t$ \\
    Actual dispatched wind energy & Amount of wind energy actually dispatched at time $t$. & $x_{wt} = b_{wt}$ if $p_{wt} \leq c_t$ \newline otherwise $0$ \\
    Wind-power revenue & Revenue of the wind-power producer at time $t$. Here, $b_{wt}$ and $p_{wt}$ are the bid quantity and price of the wind-power producer; $u_t = max(0, x_{wt}-G_t)$. & $f(revenue) = c_tx_{wt} - q_uu_t$ \\

   \bottomrule
  \end{tabular*}
  \caption{Time series of variables used to evaluate the consistency in the performance of final Python implementations in the electricity market problem.}
  \label{market_series}
\end{table}

\begin{table}
  \begin{tabular*}{\tblwidth}{ >{\raggedright\arraybackslash}p{3cm}
                               >{\raggedright\arraybackslash}p{8.6cm}
                               >{\raggedright\arraybackslash}p{3.9cm} }
   \toprule
    Parameters & Descriptions & Values (dimensionless) \\
   \midrule
    $b$ & Pollution removal rate & 0.42 \\
    $q$ & Recycling exponent & 2.0 \\
    $\mu$ & Mean of natural pollution inflows & 0.02 \\
    $\sigma$ & Standard deviation of natural pollution inflows & 0.0017 \\
    $\delta$ & Discount factor & 0.98 \\
    $a_{t}$ & Anthropogenic pollution emission at time $t$ & Fixed sequence uniformly sampled from [0.02, 0.04] \\
    $r_{t}$ & Regulator’s pollution removal at time $t$ & [0, 0.001, 0.002, 0.003] \\
    $seed$ & Random seed & 4521 \\
   \bottomrule
  \end{tabular*}
  \caption{Experimental settings used to generate the time series in the lake problem.}
  \label{lake_parameter}
\end{table}

\begin{table}
  \begin{tabular*}{\tblwidth}{ >{\raggedright\arraybackslash}p{3cm}
                               >{\raggedright\arraybackslash}p{8cm}
                               >{\raggedright\arraybackslash}p{5cm} }
   \toprule
    Parameters & Descriptions & Values \\
   \midrule
    $\mu_{D}$ & Mean of market demands & 800 (MWh) \\
    $\sigma_{D}$ & Standard deviation of market demands & 200 (MWh) \\
    $b_i$ & Bid quantity for the conventional producer $i$ & [300, 250, 1000] (MWh) \\
    $\mu_{pi}$ & Mean of bid prices for the conventional producer $i$ & [45, 50, 60] (\$/MWh) \\
    $\sigma_{pi}$ & Standard deviation of bid prices for the conventional producer $i$ & [5, 5, 5] (\$/MWh) \\
    $(a,b)$ & Solar-power production parameters & (0, -400) (MWh) \\
    $\mu_{ps}$ & Mean of bid prices for the solar-power producer $s$ & 35 (\$/MWh) \\
    $\sigma_{ps}$ & Standard deviation of bid prices for the solar-power producer $s$ & 10 (\$/MWh) \\
    $\mu_G$ & Mean of wind-power production & 300 (MWh) \\
    $\sigma_G$ & Standard deviation of wind-power production & 100 (MWh) \\
    $(b_{wt}, p_{wt})$ & Bid for the wind-power producer at time $t$ & (300, 50) (MWh, \$/MWh) \\
    $q_{u}$ & Shortfall penalty coefficient & [0, 20, 80, 140] (\$/MWh) \\
    $seed$ & Random seed & 4521 (dimensionless) \\
   \bottomrule
  \end{tabular*}
  \caption{Experimental settings used to generate the time series in the electricity market problem.}
  \label{market_parameter}
\end{table}

%% file: results.tex
\section{Results}
This section presents our experimental results. Detailed information can be found in our GitHub Repository \url{https://github.com/Hardy-Pei01/LLM-for-DMDU.}

\subsection{Workflow demonstration}
This section demonstrates the communication with ChatGPT 5.2 Instant following the workflow on the three sub-cases (Figure \ref{fig:experiments}). It mainly refers to the three conversation logs (Lake\textunderscore1, Market\textunderscore1 and Market\textunderscore brief\textunderscore1) in the supplementary materials.

\subsubsection{Initial formalization}
For the lake problem, we provided a comprehensive problem description covering the essential components for modeling as input (the description is presented in Section \ref{lake_problem}). Then, the LLM was prompted to formalize the problem as introduced before.

The initial response from the LLM failed to pass \emph{design concepts} in the checklist, because it incorrectly treated the eutrophication threshold $X_{crit}$ as a constant. This error was corrected after one refinement iteration. In the second response, the output model specification included the following components:
\begin{enumerate}
    \item The revised definition of the eutrophication threshold.
    \item Essential model components directly extracted from the input description, including the time index, state variables, decision variables, stochastic variables, deep uncertainties, transition function and failure condition.
    \item An inferred objective, which is to maximize expected discounted benefits while avoiding failure across deep uncertainties. This objective is inferred based on the benefit equation and the failure condition provided in the input description.
    \item Candidate robust metrics used to assess the robustness of the system.
    \item An inferred form of the simulation model.
\end{enumerate}
This model specification passed all items in the checklist (Section \ref{checklist}), so we argued that it was consistent with the input description. We also observed that the LLM not only extracted essential information from the input description (Components 1 and 2), but also complemented the specification with additional components derived from its common-sense knowledge (Components 3-5). Furthermore, all the components were explicitly explained in the specification.

For the comprehensive-description electricity market problem, the input description is presented in Section \ref{market_problem}. The initial response from the LLM failed to pass \emph{design concepts}, because it did not explicitly specify that the means and variances of the market demand and the actual wind-power production are constants. Therefore, we issued a refinement prompt to correct that. The final model specification was consistent and included the following components:
\begin{enumerate}
    \item Essential model components directly extracted from the input description, including the time index, market structure, constants, deep uncertainties, stochastic variables and decision variables.
    \item Candidate state variables, including the market demand, actual wind-power production, and the bid prices of the conventional and solar-power producers at the current dispatch interval.
    \item A mathematical expression of the market-clearing mechanism, constructed based on the input description and the LLM’s common-sense knowledge.
    \item An inferred objective, which is to maximize the expected daily revenue of the wind-power producer, accounting for the shortfall penalty.
    \item An inferred form of the simulation model.
\end{enumerate}
Similarly, this specification also included the extracted information (Component 1) and complemented information (Component 2-5). All these components were explicitly explained.

For the brief-description electricity market problem, the input description is as follows:
\begin{tcolorbox}[enhanced, colback=blue!5!white, colframe=blue!75!black]
  
  An energy company is interested in investing in wind-power production. To assess the viability of this investment, the company wants to evaluate the expected revenue of a wind-power producer in a competitive spot market. \\
  
  Suppose there are five energy producers participating in the day-ahead spot market. Three of them use conventional sources for energy production, such as coal, while the other two producers use solar and wind power, respectively. The market is cleared once per day for 24 hourly intervals of the following day. For each dispatch interval, every producer submits a bid specifying the energy quantity (MWh) it is willing to supply and the corresponding minimum acceptable price (per MWh). Then, the market operator follows a merit-order process to determine the market-clearing price and dispatch schedule, ensuring the total market demand is met. If a bid is accepted, the producer is committed to delivering the bid quantity at the market-clearing price. Otherwise, the producer earns zero revenue. \\
  
  The energy company aims to build a simulation model to evaluate the expected revenue of the wind-power producer on any future day by selecting the hourly bid quantity and price, while accounting for uncertainties. This model can then be used to explore outcomes across various short-term and long-term scenarios, thereby informing the viability of the investment.
\end{tcolorbox}
This description only briefly describes the market structure, market-clearing mechanism and the aim of this problem. Unsurprisingly, the initial specification from the LLM lacked several details, including (1) the function for computing the market demand; (2) the behavioral logic of the conventional, solar-power and wind-power producers; and (3) the shortfall penalty mechanism for the wind-power producer. Therefore, the specification failed to pass \emph{design concepts} and \emph{processes and scheduling} in the checklist. To complement the specification, we issued three prompts to provide information related to (1), (2) and (3) respectively. However, the refined specification still failed to pass \emph{design concepts} because it considered deep uncertainties arising from long-term shifts in some constants. After another refinement, the LLM produced the final specification consistent with the baseline description, which included:
\begin{enumerate}
    \item A modeling framework overview, which summarizes the model type, purpose of the problem, time index, and a potential solution approach.
    \item The rest of this specification covers Components 1–4 of the specification in the comprehensive-description electricity market problem, but with slight differences.
\end{enumerate}
This also indicates that the LLM has formed a comprehensive understanding of the problem by the end of Step 1. Consequently, the performance of the LLM in the brief-description electricity market problem would be similar to that in the comprehensive-description electricity market problem in the subsequent steps of the workflow.

All the sub-cases showed that the LLM was able to extract relevant information from the description, complement the specification with its common-sense knowledge, and explicitly explain these components. This improved the readability of the output specification and facilitated researchers’ understanding of the problem. These observations indicate that the use of LLMs could effectively support the problem conceptualization process.

\subsubsection{Multi-perspective formalization}
After constructing the initial model specification, the LLM was prompted to refine the specification by modeling two pre-defined perspectives at Step 2. We used a fixed number of pre-defined perspectives here to ensure that the LLM outputs across different runs were generally consistent and comparable.

In the lake problem, the two perspectives were local community and environmental regulator. The output model had the following structure: 
\begin{enumerate}
    \item Common environment
    \begin{enumerate}
        \item This layer specifies the components common to both perspectives, including:
        \begin{enumerate}
            \item Global components from Step 1, including the time index, state variables, stochastic variables, deep uncertainties and eutrophication threshold.
            \item A general transition function, where the decision variable is only broadly defined as an aggregation of decision variables from all perspectives without an explicit definition.
        \end{enumerate}
    \end{enumerate}
    \item Perspective 1: local community
    \begin{enumerate}
        \item It represents industry and agriculture seeking economic benefits from pollution-generating activities. The perspective-specific components include:
        \begin{enumerate}
            \item A decision variable, which is the annual anthropogenic emission.
            \item Other components, including the available information, community-controlled transition and community objectives.
        \end{enumerate}
    \end{enumerate}
    \item Perspective 2: environmental regulator
    \begin{enumerate}
        \item It represents a public authority tasked with protecting the lake, potentially at the cost of reduced economic output. The perspective-specific components include:
        \begin{enumerate}
                \item Three candidate decision variables: which are emission cap, pollution removal and pollution tax.
                \item Other components, including the regulator-controlled transition and regulator objectives.
        \end{enumerate}
    \end{enumerate}
\end{enumerate}

In both sub-cases of the electricity market problem, the two perspectives were wind-power producer and system regulator. Their output models were generally similar and typically had the following structure: 
\begin{enumerate}
    \item Common environment
    \begin{enumerate}
        \item This layer specifies the components common to both perspectives, including:
        \begin{enumerate}
            \item Global components from Step 1, including the time index, market structure, constants, deep uncertainties, and stochastic variables.
            \item The market-clearing mechanism.
        \end{enumerate}
    \end{enumerate}
    \item Perspective 1: wind-power producer
    \begin{enumerate}
        \item It represents a profit-maximizing agent exposed to shortfall penalties. The perspective-specific components include:
        \begin{enumerate}
            \item Two decision variables, which are the bid quantity and price for hourly intervals.
            \item Other components, including the producer transition components and producer objectives.
        \end{enumerate}
    \end{enumerate}
    \item Perspective 2: system regulator
    \begin{enumerate}
        \item It represents regulatory bodies who do not bid energy, but set market rules to ensure system reliability and economic efficiency. The perspective-specific components include:
        \begin{enumerate}
                \item A decision variable, which is the shortfall penalty coefficient (Other options exist, but they are not explicitly defined).
                \item Other components, including the regulator transition components and regulator objectives.
        \end{enumerate}
    \end{enumerate}
\end{enumerate}
Although the only additional information provided in the prompt at this step was the names of the two perspectives, the LLM was still able to complement the multi-perspective specification and explain the additional components without any refinement. This observation suggested that LLMs may also be effective in supporting more complex cases. For example, in real-world socio-environmental planning applications involving various stakeholder groups, it may not be feasible to engage all stakeholders in the decision-making process, potentially leading to overlooked perspectives. In these cases, LLMs can support brainstorming and defining relevant factors and perspectives that might otherwise be neglected, using their common-sense knowledge. This further demonstrated how LLMs can facilitate the problem conceptualization process.

It is also worth noting that, at this step, the output models did not yet compose the two perspectives. For example, in the lake problem, the common environment layer only specified a general transition function where the aggregated decision variable was not explicitly defined. In addition, the perspective layers were specified independently of one another. We leave this task for the next step so that this workflow can be applied to more complex cases. In those cases, more candidate perspectives and corresponding decision variables are explored at this step, making their simultaneous composition challenging and more prone to errors. Therefore, only exploration is performed at this step. After that, researchers can select the perspectives and decision variables they find appropriate and proceed to the composition step.

\subsubsection{Composition}
After generating the specification regarding multiple perspectives, the LLM was prompted to compose all components into a unified model in Step 3. The resulting model shared a structure similar to that in Step 2, but differed in the following aspects: (1) the general transition function in the common environment layer was explicitly defined as a function of the decision variables from each perspective; (2) the decision variables from each perspective were defined as exogenous inputs to the other perspectives and (3) the definitions of perspective-specific transition components and objective functions explicitly accounted for the influence of these exogenous inputs. In this way, interactions between the perspectives were described at both the common-environment level and the perspective level. For example, in the lake problem, the aggregated decision variable was defined, in both the general transition function and all perspective-specific transition components, as 
\begin{equation}
    Anthropogenic\ emission\ (community's\ action) - Pollution\ removal\ (regulator's\ action)
\end{equation}
Readers can refer to the conversation logs for more details. In addition, no refinement was required at this step in any of the sub-cases.

\subsubsection{Python implementation}
Eventually, the LLM was prompted to generate a modular Python implementation of the unified model. In all the sub-cases, the output implementation shared the same structure:
\begin{itemize}
    \item Class 1: common environment
    \item Class 2: perspective 1
    \item Class 3: perspective 2
    \item Function 1: simulation interface function
\end{itemize}
We validated the implementation using the tests in Table \ref{lake_tests} and \ref{market_tests}. In the lake problem, the implementation failed the \emph{unit tests}. Three errors were identified. First, the implementation attempted to read a deeply uncertain parameter from the dictionary of constants. Second, a grid search was used to estimate the eutrophication threshold. This method is computationally inefficient and yields low accuracy. Additionally, due to inappropriate choices of the upper and lower bounds in the implementation, the grid search failed to produce the correct result in the default scenario. Third, the log-normal distribution for natural pollution inflows was implemented incorrectly. We issued three refinement prompts to ask the LLM to fix the errors. After that, the final implementation passed all tests. In both sub-cases of the electricity market problem, the output implementations failed the \emph{property-based test} because bids priced at the clearing price were not fully accepted when the aggregate accepted supply exceeded the market demand. After issuing one corrective prompt, the final implementations passed all tests. This observation indicates that the LLM is prone to errors when implementing complex functionalities, such as root-finding algorithms, log-normal distributions, and the merit-order market-clearing mechanism. Researchers should exercise increased care when applying the proposed workflow for problem conceptualization.

\begin{table}
  \begin{tabular*}{\tblwidth}{ >{\raggedright\arraybackslash}p{3cm}
                               >{\raggedright\arraybackslash}p{13cm}}
   \toprule
    Tests & Details \\
   \midrule
    Unit tests & They verified the correctness of the log-normal distribution sampling and the calculation of the eutrophication threshold. \\
    Property-based test & It verified whether lake eutrophication is irreversible in the absence of regulatory intervention. \\
    Scenario test & It verified whether the program's performance was consistent with the standard implementation from EMA Workbench \citep{kwakkel2017exploratory} in a specific scenario (described in Section \ref{scenario}). \\
   \bottomrule
  \end{tabular*}
  \caption{Tests used for verification in \emph{Python implementation} in the lake problem.}
  \label{lake_tests}
\end{table}

\begin{table}
  \begin{tabular*}{\tblwidth}{ >{\raggedright\arraybackslash}p{3cm}
                               >{\raggedright\arraybackslash}p{13cm}}
   \toprule
    Tests & Details \\
   \midrule
    Unit tests & They verified that the profit and penalty calculations for the wind-power producer were correct and that total offered supply was sufficient to meet the market demand. \\
    Property-based test & It verified the correctness of the implementation of the merit-order market-clearing mechanism. \\
    Scenario test & It verified whether the program's performance was consistent with the baseline implementation in a specific scenario (described in Section \ref{scenario}). \\
   \bottomrule
  \end{tabular*}
  \caption{Tests used for verification in \emph{Python implementation} in the electricity market problem.}
  \label{market_tests}
\end{table}

\subsection{Components extraction consistency}
In every sub-case, the LLM was able to correctly extract all key components (specified in Table \ref{lake_components} and \ref{market_components}) by the end of Step 1, guided by the initial prompt and subsequent iterative refinements (Table \ref{components_results}). We also observed that these components were preserved with minor errors in subsequent responses throughout the conversation. Although the LLM performed poorly at the beginning in the brief-description electricity market problem, it was because the initial prompt provided only a brief narrative, necessitating additional refinements to complement the picture of the baseline problem. Therefore, we argued that these results indicate that the LLM was highly consistent in components extraction when following the workflow.

\begin{table}
  \begin{tabular*}{\tblwidth}{ >{\raggedright\arraybackslash}p{2.3cm}
                               >{\centering\arraybackslash}p{1.3cm}
                               >{\centering\arraybackslash}p{1.4cm}
                               >{\centering\arraybackslash}p{1.5cm}
                               >{\centering\arraybackslash}p{1.5cm}
                               >{\centering\arraybackslash}p{1.6cm}
                               >{\centering\arraybackslash}p{1.8cm}
                               >{\centering\arraybackslash}p{1.4cm}
                               >{\centering\arraybackslash}p{2cm}}
   \toprule
    & State variables & Decision variables & Transition function & Stochastic variables & Constants & Deep uncertainties & Objective function & Other component \\
   \midrule
    Lake (before) & 1 & 1 & 1 & 1 & 1 & 5 & 1 & 0.25 \\
    Lake (after) & 1 & 1 & 1 & 1 & 1 & 5 & 1 & 1 \\
    Comprehensive market (before) & 0 & 2 & 1 & 6 & 9 & 8 & 1 & 1 \\
    Brief market (before) & 0 & 2 & 1 & 5 & 0 & 0 & 0 & 0 \\
    Market (after, for both) & 0 & 2 & 1 & 6 & 10 & 8 & 1 & 1 \\
   \bottomrule
  \end{tabular*}
  \caption{Number of components correctly extracted by the LLM before and after refinements in Step 1. These numbers were averaged over four runs in each sub-case. Results for Step 2-4 were not reported because all these components were correctly identified by the end of Step 1, and were preserved with minor errors in subsequent steps. The two sub-cases of the electricity market problem share the same “after” results because they are based on the same baseline description.}
  \label{components_results}
\end{table}

\subsection{Model formalization consistency}
In all sub-cases, the LLM was able to produce consistent responses after only a small number of iterations when following our workflow (Table \ref{iterations}). Generally, the LLM performed best in \emph{composition}, where no refinement was required. In the lake problem and the comprehensive-description electricity market problem, the LLM performed particularly poorly in \emph{Python implementation}. The main factor was that the LLM struggled to correctly implement some complex functionalities without explicit specifications of their implementation details. In the lake problem, these functionalities included the computation of the eutrophication threshold and the implementation of the log-normal distribution. In the electricity market problem, this functionality was the implementation of the merit-order market-clearing mechanism. In the brief-description electricity market problem, the LLM performed worst in \emph{initial formalization}. This was because the input narrative was brief and omitted much of the necessary information, which was added through subsequent iterative refinements. Once the LLM completed \emph{initial formalization}, the performance of the LLM was similar to that in the comprehensive-description electricity market problem in the subsequent steps. In all sub-cases, the LLM also made errors occasionally in other steps, which were easily corrected. These results indicate that the LLM maintained consistency in model formalization when following the workflow, but human validation and refinement were still necessary.

\begin{table}
  \begin{tabular*}{\tblwidth}{ >{\raggedright\arraybackslash}p{4cm}
                               >{\raggedright\arraybackslash}p{3.6cm}
                               >{\raggedright\arraybackslash}p{4.1cm}
                               >{\raggedright\arraybackslash}p{4.1cm}}
   \toprule
    & Lake Problem \newline [mean $\pm$ std] & Electricity Market Problem (comprehensive) \newline [mean $\pm$ std] & Electricity Market Problem (brief) [mean $\pm$ std] \\
   \midrule
    Initial formalization & $0.8 \pm 0.4$ & $0.8 \pm 0.4$ & $4.5 \pm 0.5$ \\
    Multi-perspective formalization & $0.3 \pm 0.4$ & $0.5 \pm 0.5$ & $0.3 \pm 0.4$ \\
    Composition & $0.0 \pm 0.0$ & $0.0 \pm 0.0$ & $0.0 \pm 0.0$ \\
    Python implementation & $2.5 \pm 0.5$ & $1.5 \pm 0.9$ & $1.3 \pm 0.4$ \\
   \bottomrule
  \end{tabular*}
  \caption{Average number of iterations required by the LLM to produce a consistent response at each step in each sub-case. These numbers were averaged over four runs.}
  \label{iterations}
\end{table}

\subsection{Python implementation consistency}
\begin{figure}
    \centering
    \begin{subfigure}{.45\textwidth}
        \centering
        \captionsetup{width=.9\linewidth}
        \includegraphics[width=\linewidth]{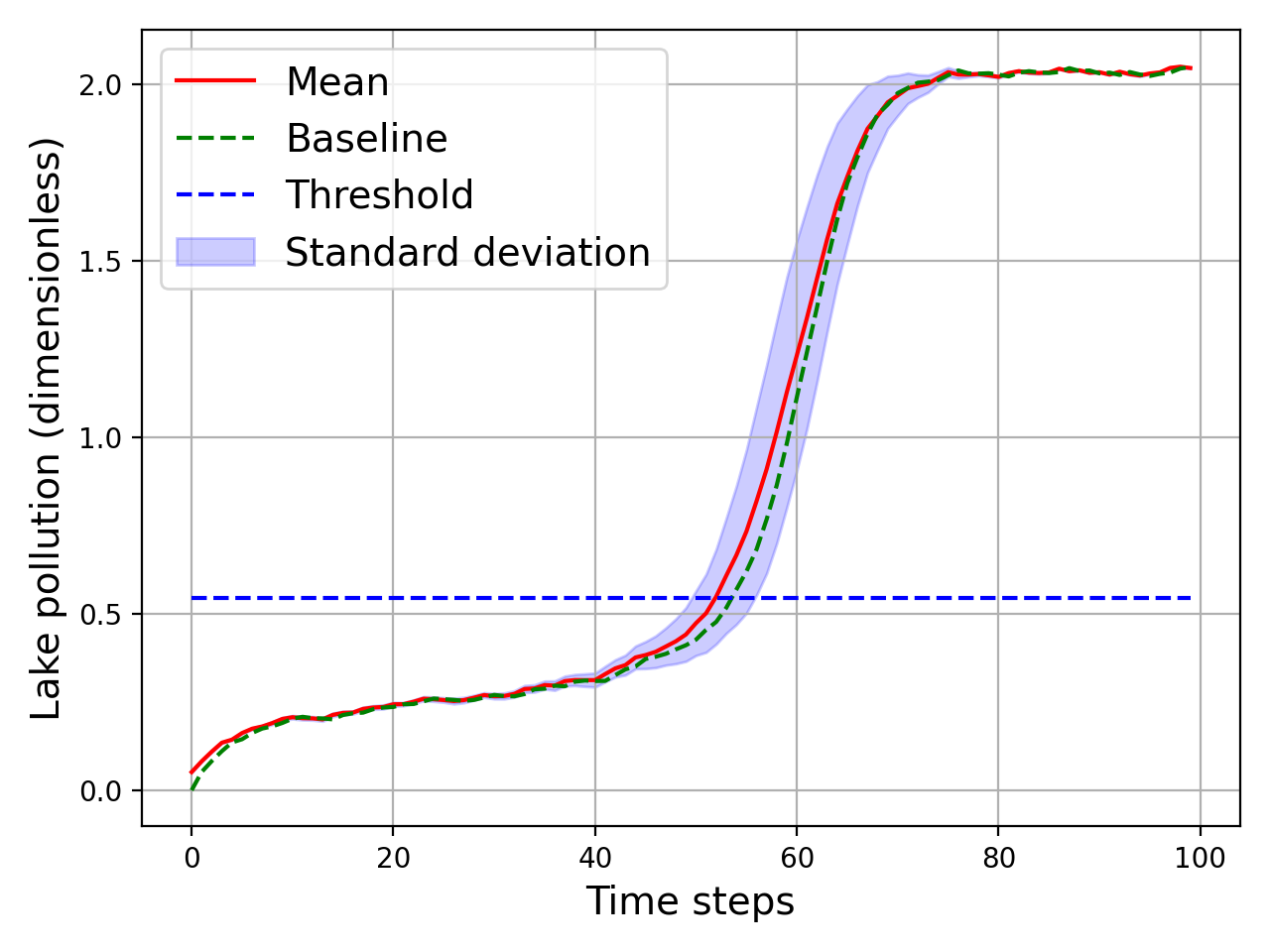}
        \caption{Lake pollution with $\bar{a_t}$ and $r_t = 0$.}
        \label{fig:lake1}
    \end{subfigure}
    \begin{subfigure}{.45\textwidth}
        \centering
        \captionsetup{width=.9\linewidth}
        \includegraphics[width=\linewidth]{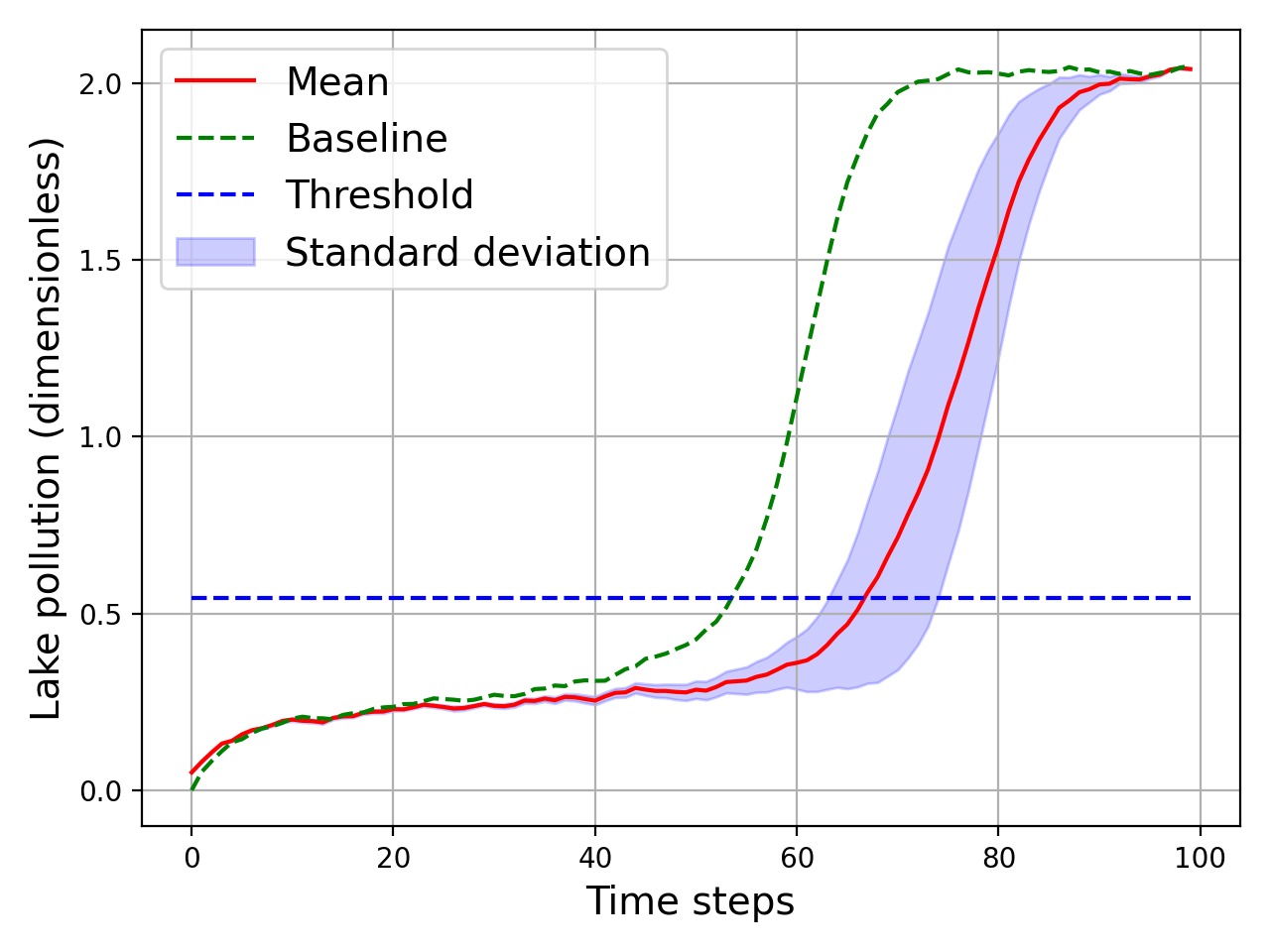}
        \caption{Lake pollution with $\bar{a_t}$ and $r_t = 0.001$.}
        \label{fig:lake2}
    \end{subfigure}
    \begin{subfigure}{.45\textwidth}
        \centering
        \captionsetup{width=.9\linewidth}
        \includegraphics[width=\linewidth]{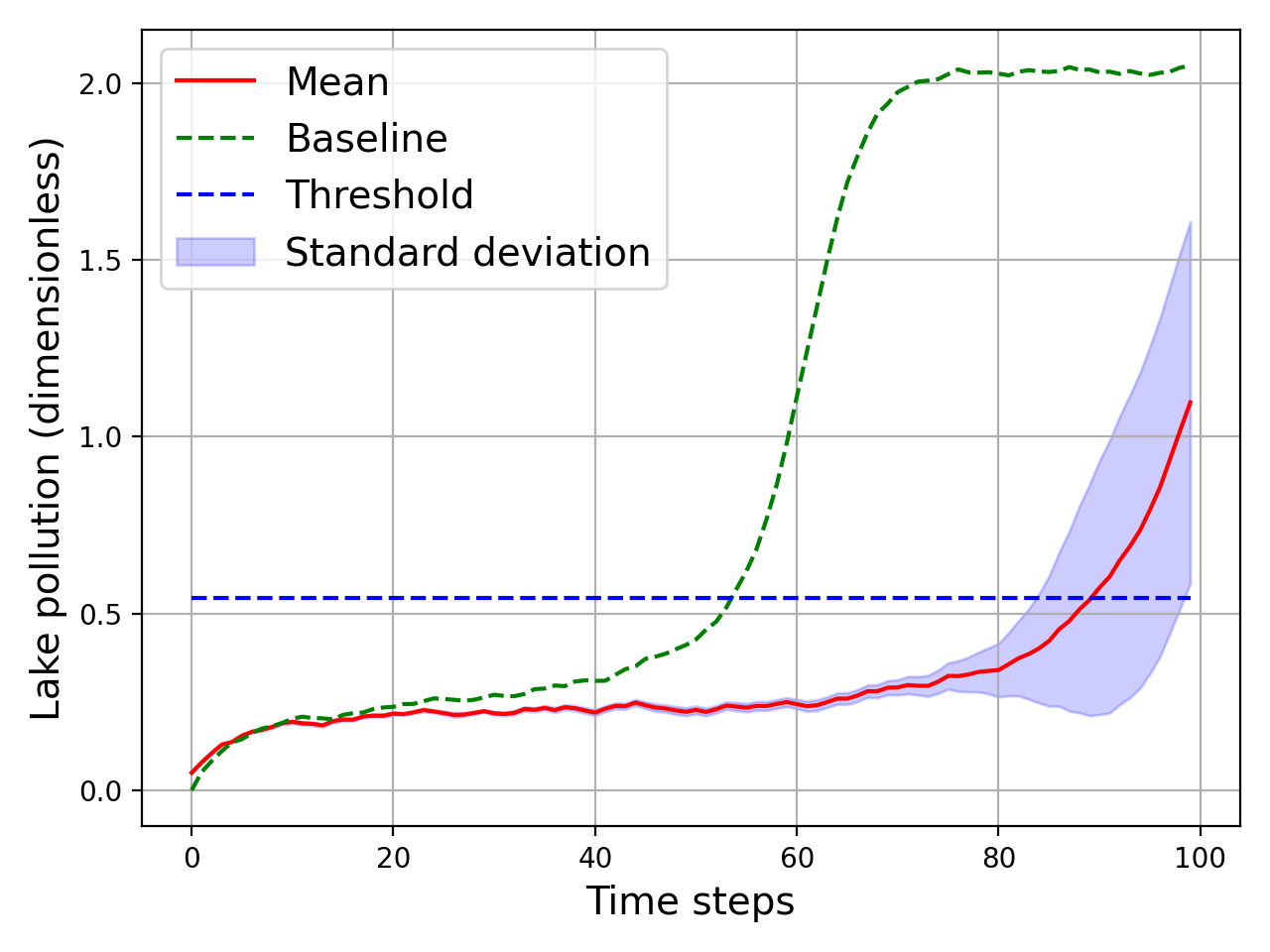}
        \caption{Lake pollution with $\bar{a_t}$ and $r_t = 0.002$.}
        \label{fig:lake3}
    \end{subfigure}
    \begin{subfigure}{.45\textwidth}
        \centering
        \captionsetup{width=.9\linewidth}
        \includegraphics[width=\linewidth]{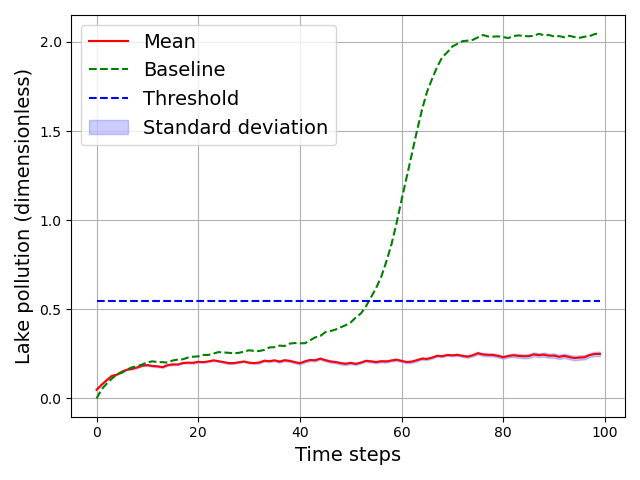}
        \caption{Lake pollution with $\bar{a_t}$ and $r_t = 0.003$.}
        \label{fig:lake4}
    \end{subfigure}
    \caption{Average time series of lake pollution for both perspectives, with a fixed sequence of pollution emission $\bar{a_t}$ (decisions from Perspective 1: local community), and varying pollution removal levels $r_t$ (decisions from Perspective 2: environmental regulator). To generate each of these figures, we ran the Python implementation produced in each experimental run ten times, collecting a total of 40 time series. Figure \ref{fig:lake1} shows the impact of Perspective 1 on the system alone, while Figure \ref{fig:lake2}–\ref{fig:lake4} show the combined impact of Perspective 1 and the increasing influence of Perspective 2 on the system.}
    \label{fig:lake_results}
\end{figure}

Figure \ref{fig:lake_results} shows the average time series of lake pollution produced by the output Python implementations in our lake problem experiments, using the experimental settings in Table \ref{lake_parameter}. We also used the result generated by the standard implementation in EMA Workbench \citep{kwakkel2017exploratory} as the baseline for comparison. Figure \ref{fig:lake1} shows the results for the scenario without intervention by the environmental regulator. In this case, the problem conceptualized by the LLM is identical to the standard lake problem specified in EMA Workbench. It can be observed that the mean of our time series is close to the baseline. In fact, the observed difference was solely caused by the stochastic pollution inflow in the implementation. If this variable is held constant, our time series would be identical to the baseline. This indicates that our workflow is able to produce Python implementations that are consistent with the input problem description. Figure \ref{fig:lake2} - \ref{fig:lake4} show the results under increasing levels of intervention by the environmental regulator. As pollution removal at each time step ($r_t$) increased, lake pollution accumulated more slowly over time. When $r_t = 0.003$, the lake system reached a balance between pollution inflow and removal, thereby maintaining lake pollution below the eutrophication threshold at all times. This indicates that the workflow maintains consistency across multiple perspectives. Additionally, it is worth noting that the standard deviations of these time series are small, which were also solely caused by the stochastic pollution inflow. This indicates that the workflow maintains consistency across different experimental runs.

\begin{figure}
    \centering
    \begin{subfigure}{.45\textwidth}
        \centering
        \captionsetup{width=.9\linewidth}
        \includegraphics[width=\linewidth]{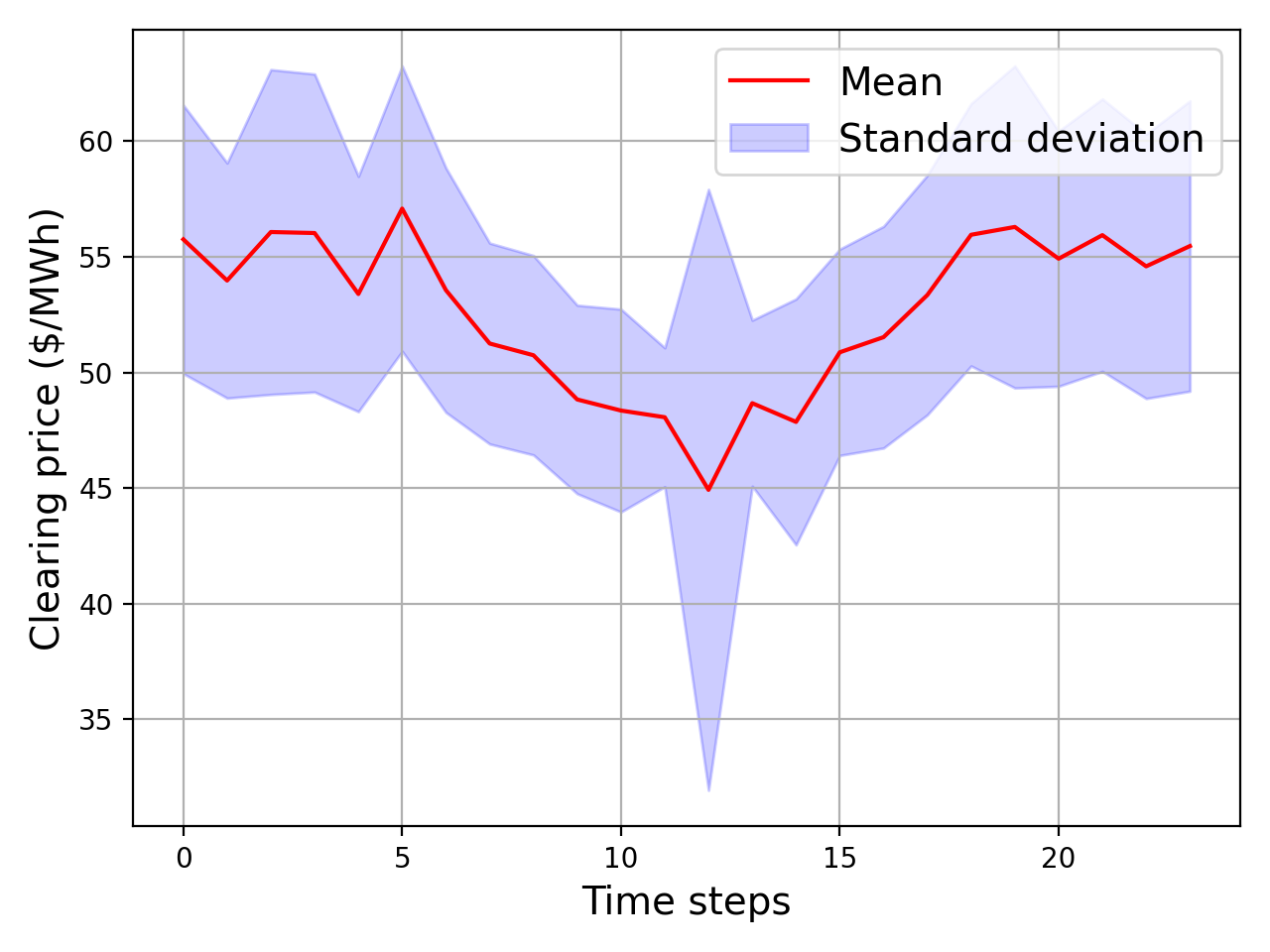}
        \caption{Clearing prices with $(\bar{b_{wt}}, \bar{p_{wt}})$ and $q_u = 0$.}
        \label{fig:market1}
    \end{subfigure}
    \begin{subfigure}{.45\textwidth}
        \centering
        \captionsetup{width=.9\linewidth}
        \includegraphics[width=\linewidth]{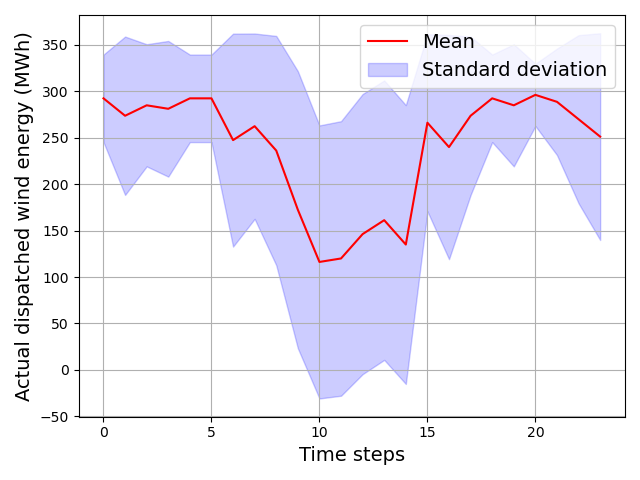}
        \caption{Actual dispatched wind energy with $(\bar{b_{wt}}, \bar{p_{wt}})$ and $q_u = 0$.}
        \label{fig:market2}
    \end{subfigure}
    \begin{subfigure}{.45\textwidth}
        \centering
        \captionsetup{width=.9\linewidth}
        \includegraphics[width=\linewidth]{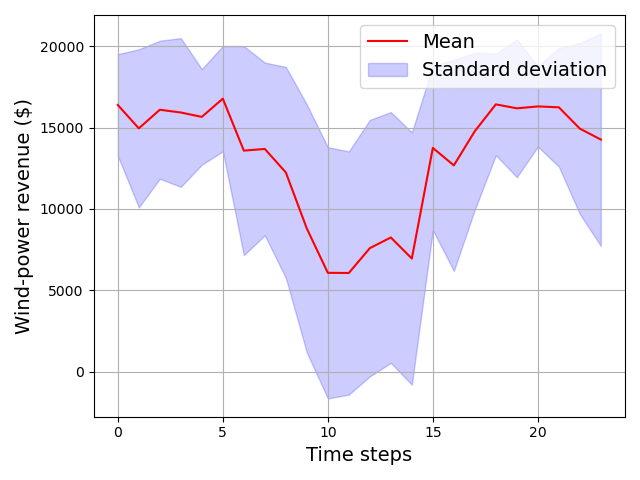}
        \caption{Wind-power revenue with $(\bar{b_{wt}}, \bar{p_{wt}})$ and $q_u = 0$.}
        \label{fig:market3}
    \end{subfigure}
    \begin{subfigure}{.45\textwidth}
        \centering
        \captionsetup{width=.9\linewidth}
        \includegraphics[width=\linewidth]{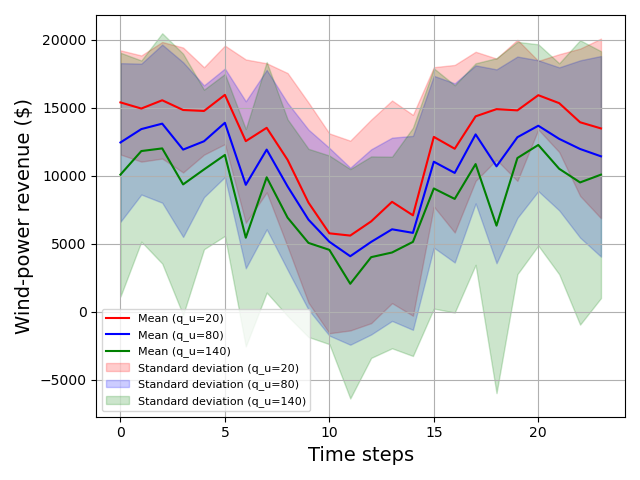}
        \caption{Wind-power revenue with $(\bar{b_{wt}}, \bar{p_{wt}})$ and varying $q_u$.}
        \label{fig:market4}
    \end{subfigure}
    \caption{Average time series of clearing prices, actual dispatched wind energy and wind-power revenue in both sub-cases of the electricity market problem for both perspectives, with a fixed sequence of bids for the wind-power producer $(\bar{b_{wt}}, \bar{p_{wt}}) = (300, 50)$ (decisions from Perspective 1: wind-power producer), and varying shortfall penalty coefficients $q_{u}$ (decisions from Perspective 2: system regulator). To generate each of these figures, we ran the Python implementation produced in each experimental run ten times, collecting a total of 80 time series. Figure \ref{fig:market1}-\ref{fig:market3} show the impact of Perspective 1 on the system alone, while Figure \ref{fig:market4} shows the combined impact of Perspective 1 and the increasing influence of Perspective 2 on the system.}
    \label{fig:market_results}
\end{figure}

Figure \ref{fig:market_results} shows the average time series of clearing price, actual dispatched wind energy and wind-power revenue produced by the output Python implementations in the two sub-cases of the electricity market problem, using the experimental settings in Table \ref{market_parameter}. The series from different sub-cases could be aggregated because the LLM was prompted to conceptualize the same baseline problem, albeit with different initial prompts. Figure \ref{fig:market1} - \ref{fig:market3} show the results for the scenario without intervention by the system regulator. It can be observed that these series exhibit consistent temporal patterns, with their values decreasing significantly around the middle of the day. This is consistent with merit-order dispatch interacting with time-varying solar-power production. According to the problem description (Section \ref{market_problem}) and parameter settings (Table \ref{market_parameter}), solar-power production has the lowest cost and peaks during the day. Consequently, it is dispatched with the highest priority, leading to a reduction in the clearing price, actual dispatched wind energy and corresponding wind-power revenue around the middle of the day. This indicates that the workflow maintains consistency with the input problem description. Figure \ref{fig:market4} shows the results under increasing levels of intervention by the system regulator. Increasing the shortfall penalty coefficient did not significantly alter the temporal pattern, because it did not affect the decisions of the wind-power producer according to the model specification. It slightly decreased the average wind-power revenue while significantly increasing its variance. This was because the penalty was only applied when under-delivery occurred, which is consistent with the multi-perspective specification. Additionally, although the standard deviations of these time series (Figure \ref{fig:market_results}) are larger than those in the lake problem (Figure \ref{fig:lake_results}), it was due to the electricity market problem involving more stochastic variables. If these variables are held constant, the time series across different experimental runs would also be identical and have no variation. This indicates that the workflow maintains consistency across multiple runs. 

Note that the two problems demonstrated here are hypothetical. They are presented solely to show their simulation capabilities, while decision-making processes and mechanisms are not considered. Specifically, decisions from each perspective are pre-defined and do not respond to decisions from the other perspective in these illustrations. For this reason, readers should not rely on these results for real-world applications.

\FloatBarrier

%% file: discussion.tex
\section{Discussion}
Based on the literature review and experiments conducted, this section provides a balanced assessment of the strengths and weaknesses for applying the proposed LLM-assisted workflow to facilitate problem conceptualization in socio-environmental planning. 

\subsection{Strengths of the LLM-assisted workflow}
A primary strength of the workflow is its potential to \textbf{improve the efficiency of problem conceptualization}. By leveraging LLMs' exceptional capabilities in understanding and processing natural language, researchers can translate an intuitive problem narrative into a structured model specification within only a few iterations following the workflow. This will reduce the time and lower the knowledge barrier required to form an initial conceptual representation of the problem, thereby encouraging broader stakeholder participation in the planning process. In real-world socio-environmental planning contexts, the problem conceptualization process often involves extensive communication, interpretation and negotiation among researchers, decision-makers and other stakeholders. The use of this workflow would allow them to shift effort from manual transcription to higher-value conceptualization and adjudication.

Another strength of the workflow is its support for \textbf{breadth exploration}. Real-world socio-environmental planning applications often involve various groups of stakeholders with different perceptions, backgrounds, and interests, and they may affect or be affected by the decisions in different ways. Since it is often unfeasible or inefficient to involve all stakeholders in the planning process, it is challenging for researchers to take all relevant perspectives into consideration during problem conceptualization. Our workflow demonstrates how LLMs can be used to explore alternative scenarios and perspectives, along with their corresponding decisions and objectives, to overcome this challenge. Moreover, this capability also aligns with the idea of exploratory analysis in DMDU that explores all possibilities for robust decision-making.

This workflow also bridges conceptual and computational problem presentations by supporting \textbf{programmatic implementation}. As demonstrated, LLMs can be used to translate the model specification into an executable Python program, even when the user is not an expert programmer. In this way, LLMs facilitate the programming process, and further reduce the technical barrier to stakeholder participation to some extent, given that the output program is not guaranteed to be error-free. 

Eventually, the workflow provides \textbf{traceability and consistency} to some extent. LLMs operate through persistent conversations, and the conversation log is preserved. In addition, our experiments demonstrate that LLMs are able to maintain general consistency across the conversation. This capability makes the problem conceptualization process more transparent and reliable, allowing researchers to trace and advance progress more efficiently. This capability is particularly important in socio-environmental planning applications, since they often involve long time horizons.

\subsection{Limitations and future research}
In addition to the strengths, the workflow also inherits the limitations of LLMs. For this reason, these limitations may also arise across LLM-based applications. This section discusses these limitations and provides potential strategies to mitigate them. Furthermore, this discussion also opens up directions for future research on LLM applications.

First, the workflow is subject to \textbf{limited long-horizon reliability}, since LLMs cannot guarantee their performance in long-horizon reasoning \citep{momennejad2023evaluating, pallagani2023understanding, valmeekam2024llms}. In complex socio-environmental planning involving intricate system dynamics, numerous stakeholders, and a large number of interacting components, the problem conceptualization process often requires many steps, and the workflow cannot be directly applied. Instead, researchers should decompose the problem into several simpler subproblems, apply the workflow incrementally, and conduct additional validation and verification to ensure the correctness of the outputs.

Second, the workflow is subject to \textbf{prompt sensitivity}. Small changes in the input prompt may lead to significant differences in the response of LLMs, particularly at the early stages of a conversation \citep{cao2024worst}. This issue can be mitigated by lowering sampling temperature, which is a hyperparameter that controls sampling randomness in LLMs \citep{renze2024effect}.

Third, the workflow is subject to \textbf{limited reproducibility}, even when identical prompts are used. This limitation arises because LLM versions are updated frequently, and different random seeds or decoding settings may be used for different conversations. Researchers may address this issue by running the workflow through the OpenAI API rather than ChatGPT and fixing all relevant settings. 

Eventually, the workflow does \textbf{not guarantee code correctness}. As demonstrated in our experiments, the Python implementations generated by the LLM are prone to logical errors. This means, although the code runs without crashing, it can still produce incorrect results due to some underlying flaws in the implemented logic. Therefore, LLMs can only be used as assistive tools for programming rather than as replacements for human programmers. Researchers may mitigate this issue by using testing to validate the output Python implementations, as we demonstrated in Table \ref{lake_tests} and \ref{market_tests}.

In summary, the proposed LLM-assisted workflow should function as a decision-support tool to facilitate socio-environmental planning, instead of completing the entire problem conceptualization process alone. As shown in Figure \ref{fig:DMDU_after}, this workflow is used for initial conceptualization, after which researchers should validate, verify and finalize the model specification and programmatic implementation based on their expert knowledge. With an appropriate use, this workflow can facilitate (1) the translation of stakeholders' narratives to structured model specifications; (2) the exploration of alternative perspectives and their corresponding decisions and objectives; and (3) the reduction of barriers to stakeholder participation.

%% file: conclusion.tex
\section{Conclusion}
Problem conceptualization is the first step in the socio-environmental planning process. It requires researchers to collaborate with decision-makers and other stakeholders to translate their qualitative problem understandings into a quantitative model to support subsequent exploratory analysis and plan development. In practice, this step can be complex and time-intensive, as stakeholders typically express their ideas in natural language, thereby requiring significant manual effort for interpretation and analysis. 

To facilitate this step, this paper proposed a templated, LLM-assisted workflow for supporting initial problem conceptualization in socio-environmental planning under deep uncertainty. By following this workflow, researchers can translate stakeholders' intuitive knowledge of the problem into an initial, model-ready specification efficiently, enabling faster transition from narrative framing to computational representation. We demonstrated the workflow using two DMDU problems and found that the LLM was able to produce satisfactory outputs and maintain consistency when following the workflow, requiring only a small number of refinement iterations. These findings demonstrate the effectiveness of the workflow. They also suggest a promising direction for future research, which is to apply and evaluate this workflow in more complex, real-world contexts that involve multiple actors and more complex systems, where this tool is crucial for supporting efficient planning processes.

%% file: ack.tex
\section*{CRediT authorship contribution statement}

\textbf{Zhihao Pei:} Conceptualization, Data curation, Formal analysis, Investigation, Methodology, Validation, Visualization, Writing – original draft; Writing – review and editing. \textbf{Nir Lipovetzky:} Conceptualization, Methodology, Supervision, Writing – review \& editing. \textbf{Angela M. Rojas-Arevalo:} Conceptualization, Methodology, Supervision, Writing – review \& editing. \textbf{Fjalar J. de Haan:} Conceptualization, Methodology, Supervision, Writing – review \& editing. \textbf{Enayat A. Moallemi:} Conceptualization, Methodology, Supervision, Writing – review \& editing. 

\section*{Declaration of Competing interest}
The authors declare that they have no known competing financial interests or personal relationships that could have appeared to influence the work reported in this paper.

\section*{Declaration of generative AI and AI-assisted technologies in the manuscript preparation process}
During the preparation of this work the author Zhihao Pei used ChatGPT in order to assist with rephrasing expressions. After using this tool/service, the author reviewed and edited the content as needed and takes full responsibility for the content of the published article.

\section*{Data Availability}
The data presented are available in our GitHub Repository https://github.com/Hardy-Pei01/LLM-for-DMDU.